\pdfoutput=1
\documentclass[11pt]{article}

\usepackage[]{acl}

\usepackage{times}
\usepackage{latexsym}

\usepackage[T1]{fontenc}
\usepackage[utf8]{inputenc}

\usepackage{microtype}

\usepackage{inconsolata}

\usepackage{dsfont}
\usepackage{bm}
\usepackage{bbm}
\usepackage{graphicx}
\usepackage{color}
\usepackage{multicol}
\usepackage{multirow}
\usepackage{wrapfig,lipsum,booktabs}
\usepackage[ruled,vlined,linesnumbered]{algorithm2e}
\usepackage{pgfplots}
\pgfplotsset{compat=1.12} 
\usepackage{xcolor}
\usepackage{colortbl}
\usepackage{tikz}
\usepackage{xspace}
\usepackage{makecell}
\usepackage{soul}
\usepackage{amsmath,amsfonts,amssymb}
\usepackage{subcaption}
\usetikzlibrary{calc}
\usepgfplotslibrary{groupplots}
\usetikzlibrary{angles,quotes} 
\usetikzlibrary{shapes,arrows}
\usetikzlibrary{backgrounds}
\usetikzlibrary{matrix}
\usetikzlibrary{patterns}
\usepackage{tikz-3dplot}
\usepackage{hyperref}
\usepackage{cleveref}
\usepackage{paralist}
\usepackage{cancel}
\usepackage{xspace}
\usepackage{todonotes}
\usepackage{tabu}
\usepackage{rotating}
\usepackage{etoolbox}
\usepackage{adjustbox}
\usepackage{enumerate}
\usepackage{enumitem}
\setitemize{noitemsep,topsep=0pt,parsep=0pt,partopsep=0pt}
\setenumerate{noitemsep,topsep=0pt,parsep=0pt,partopsep=0pt}
\usepackage{pifont}
\usepackage{cancel}
\usepackage{lipsum}
\usepackage{listings,lstautogobble}
\usepackage{fancyvrb}
\usepackage{fvextra}
\usepackage{caption}

\newcommand{\model}[1]{\textsc{#1}\xspace}
\newcommand{\docnmt}{\model{DocMT}}
\newcommand{\sennmt}{\model{SenMT}}

\newcommand{\chatgpt}{\model{gpt-3.5-turbo}}
\newcommand{\gptfour}{\model{gpt-4-turbo}}

\newcommand{\nllb}{\model{NLLB}}
\newcommand{\nllbsmall}{\model{NLLB-600M}}
\newcommand{\nllbmedium}{\model{NLLB-1.3B}}
\newcommand{\nllbbig}{\model{NLLB-3.3B}}
\newcommand{\googletrans}{\model{GoogleTrans}}

\newcommand{\mtfive}{\model{mT5}}
\newcommand{\docnmtmtfive}{\model{Doc2Doc-mT5}}
\newcommand{\docnmtmtfivesmall}{\model{Doc2Doc-mT5-300M}}
\newcommand{\docnmtmtfivebase}{\model{Doc2Doc-mT5-580M}}
\newcommand{\docnmtmtfivelarge}{\model{Doc2Doc-mT5-1.2B}}
\newcommand{\mrdoctosent}{\model{MR-Doc2Sen-mT5}}
\newcommand{\mrdoctodoc}{\model{MR-Doc2Doc-mT5}}
\newcommand{\docflat}{\model{DocFlat-mT5}}
\newcommand{\iada}{\model{IADA-mT5}}

\newcommand{\oursllamasmallfft}{\model{L-7B-fft}}
\newcommand{\oursllamasmalllora}{\model{L-7B-LoRA}}
\newcommand{\oursbloomsmallfft}{\model{B-7B-fft}}
\newcommand{\oursbloomsmalllora}{\model{B-7B-LoRA}}
\newcommand{\oursvicunasmallfft}{\model{V-7B-fft}}
\newcommand{\oursvicunasmalllora}{\model{V-7B-LoRA}}

\newcommand{\sbleu}{$s$\model{BLEU}}
\newcommand{\comet}{\model{COMET}}

\newcommand{\dbleu}{$d$\model{BLEU}}

\newcommand{\avgsbleu}{$\mu_{s\model{BLEU}}$\xspace}
\newcommand{\avgdbleu}{$\mu_{d\model{BLEU}}$\xspace}
\newcommand{\avgcomet}{$\mu_{\model{COMET}}$\xspace}

\newcommand*{\affmark}[1][*]{\textsuperscript{#1}}

\title{Adapting Large Language Models for Document-Level Machine Translation}

\author{
  Minghao Wu\affmark[$\heartsuit$]\enskip  Thuy-Trang Vu\affmark[$\heartsuit$]\enskip Lizhen Qu\affmark[$\heartsuit$]\enskip George Foster\affmark[$\spadesuit$]\enskip Gholamreza Haffari\affmark[$\heartsuit$] \\
  \affmark[$\heartsuit$]Monash University\qquad \affmark[$\spadesuit$]Google Research \\
  \texttt{\{firstname.lastname\}@monash.edu} \qquad\texttt{fosterg@google.com}
}

\begin{document}

\renewcommand{\tableautorefname}{Table}
\renewcommand{\sectionautorefname}{Section}
\renewcommand{\subsectionautorefname}{Section}
\renewcommand{\subsubsectionautorefname}{Section}
\renewcommand{\figureautorefname}{Figure}
\renewcommand{\equationautorefname}{Equation}
\renewcommand{\algorithmautorefname}{Algorithm}
\newcommand{\linenoautorefname}{Line}

\maketitle
\begin{abstract}

% Large language models (LLMs) have made significant strides in various natural language processing (NLP) tasks. Recent research shows that the moderately-sized LLMs often outperform their larger counterparts after task-specific fine-tuning. In this work, we delve into the process of adapting LLMs to specialize in document-level machine translation (\docnmt) for a specific language pair. Firstly, we explore how prompt strategies affect downstream translation performance. Then, we conduct extensive experiments with two fine-tuning methods, three LLM backbones, and 18 translation tasks across nine language pairs. Our findings indicate that in some cases, these specialized models even surpass GPT-4 in translation performance, while they still significantly suffer from the \textit{off-target translation} issue in others due to the error propagation in decoding, even if they are exclusively fine-tuned on bilingual parallel documents. Furthermore, we provide an in-depth analysis of these LLMs tailored for \docnmt, exploring aspects such as translation errors, discourse phenomena, training strategy, the scaling law of parallel documents, additional evaluation on recent test sets, and zero-shot crosslingual transfer. Our findings not only shed light on the strengths and limitations of LLM-based \docnmt models but also provide a foundation for future research.

Large language models (LLMs) have significantly advanced various natural language processing (NLP) tasks. Recent research indicates that moderately-sized LLMs often outperform larger ones after task-specific fine-tuning. This study focuses on adapting LLMs for document-level machine translation (\docnmt) for specific language pairs. We first investigate the impact of prompt strategies on translation performance and then conduct extensive experiments using two fine-tuning methods, three LLM backbones, and 18 translation tasks across nine language pairs. Our results show that specialized models can sometimes surpass GPT-4 in translation performance but still face issues like \textit{off-target translation} due to error propagation in decoding. We provide an in-depth analysis of these LLMs tailored for \docnmt, examining translation errors, discourse phenomena, strategies for training and inference, the data efficiency of parallel documents, recent test set evaluations, and zero-shot crosslingual transfer. Our findings highlight the strengths and limitations of LLM-based \docnmt models and provide a foundation for future research.

% in \docnmt.
\end{abstract}

\section{Introduction}

Large language models (LLMs) demonstrate impressive proficiency in a wide range of applications \citep{DBLP:conf/nips/Ouyang0JAWMZASR22, DBLP:conf/iclr/WeiBZGYLDDL22, DBLP:conf/iclr/SanhWRBSACSRDBX22, DBLP:journals/corr/abs-2210-11416, DBLP:journals/corr/abs-2303-08774, DBLP:journals/corr/abs-2305-10403, DBLP:journals/corr/abs-2302-13971,DBLP:journals/corr/abs-2307-09288,DBLP:journals/corr/abs-2310-06825}. However, in the realm of translation tasks, only few very large models, such as \chatgpt and \gptfour, can match or surpass the performance of state-of-the-art supervised encoder-decoder models like \nllb \citep{DBLP:journals/corr/abs-2207-04672}, while they still under-perform in translating low-resource languages \citep{robinson-etal-2023-chatgpt, DBLP:journals/corr/abs-2301-08745, DBLP:journals/corr/abs-2302-09210}. Consequently, a number of recent works attempt to bridge the gap between LLMs and supervised encoder-decoder models in translation tasks \citep{DBLP:journals/corr/abs-2304-04675, DBLP:journals/corr/abs-2305-18098, DBLP:journals/corr/abs-2306-10968, moslem-etal-2023-adaptive, DBLP:journals/corr/abs-2309-11674, DBLP:journals/corr/abs-2309-04662}. Recently, research suggests that smaller, specialized models can outperform larger, general-purpose models in specific tasks \citep{DBLP:journals/corr/abs-2306-11644, DBLP:journals/corr/abs-2306-08568, DBLP:journals/corr/abs-2310-10631}. Therefore,  we explore adapting LLMs for document-level machine translation (\docnmt) in this study.

In this study, we explore the capabilities of moderately-sized Large Language Models (LLMs) with 7 billion parameters across 18 translation tasks involving nine language pairs. We conduct experiments under both \textit{bilingual} and \textit{multilingual} settings, employing Parameter-Efficient Fine-Tuning (PEFT) and Fully Fine-Tuning (FFT) techniques on three LLM backbones. Our findings reveal that, after fine-tuning, these LLMs demonstrate superior translation performance compared to state-of-the-art models, as evidenced by metrics such as \sbleu, \dbleu, and \comet. However, a significant challenge identified is the issue of \textit{off-target translations}, which persists even after exclusive fine-tuning on bilingual corpora. Our analysis attributes this high rate of off-target translations to error propagation during the inference. Additionally, we provide a comprehensive analysis of our LLM-based \docnmt models from various perspectives. This includes examining translation error distribution, discourse phenomena, training strategies, the data efficiency of parallel documents, and additional evaluations on the recent WMT2023 test sets. We also investigate zero-shot cross-lingual transfer, aiming to enhance the understanding and efficacy of LLMs in \docnmt tasks.

We present extensive empirical evidence that highlights both the superior translation capabilities and limitations of the LLM-based \docnmt models in this study, making several significant discoveries. Here are the main takeaways:
\begin{itemize}
    \item \textbf{Selective Excellence in Translation Tasks}: 
    % Our empirical evidence reveals that our moderately-sized LLMs excel, even outperforming \gptfour, in specific translation tasks. However, they completely fail in others. We find that the primary cause of these translation failures is the \textit{off-target translation} problem. When achieving similar performance, the LLM-based \docnmt models produce fewer errors.
    Our findings show that our moderately-sized LLMs outperform \gptfour in certain translation tasks, but struggle in others due to the \textit{off-target translation} issue due to error propagation in decoding. Despite this, our \docnmt models exhibit better context awareness and fewer errors, while maintaining comparable performance.
    \item \textbf{Fine-Tuning Strategies}: 
    % We find that the PEFT approach yields superior overall performance compared to the FFT approach. However, the FFT approach demonstrates better data efficiency, requiring only about $1\%$ of the total dataset to match the performance of models trained on the entire training set. In contrast, the PEFT approach requires $10\%$ of the total dataset to achieve comparable results.
    Our research indicates that the PEFT approach outperforms the FFT approach overall. However, the FFT approach shows greater data efficiency, needing only about $1\%$ of the total dataset to reach the performance level of models trained on the entire dataset. In contrast, the PEFT approach requires $10\%$ of the total dataset for comparable results.
    \item \textbf{Evaluation on Recent Test Sets}: We evaluate our models on recent test sets between English and German from WMT2023 \citep{wmt-2023}. Our empirical results show that, when the data leakage risks are mitigated, the LLM-based \docnmt models generalize better on out-of-domain text, compared to the conventional \docnmt models.
    \item \textbf{Advantage of Base LLMs for Task-Specific Supervised Fine-Tuning}: Our study shows that base LLMs, when used as the backbone for task-specific supervised fine-tuning, perform better than instruction-tuned LLMs. They demonstrate more effective zero-shot cross-lingual transfer.
    % , making them a preferable choice.
\end{itemize}

% Our main contributions in this work are:
% \begin{itemize}
%     \item We present empirical evidence showcasing the superior translation capabilities of our LLM-based \docnmt models. Our models performs on-par or even better than \gptfour on certain translation tasks. Our findings highlight their potential to build a new paradigm of machine translation research, outperforming current advanced MT systems in selective translation tasks.
%     \item Our work includes thorough analyses of our LLM-based \docnmt models from various perspectives, deepening the understanding of how these models function and their effectiveness in different scenarios.
%     \item We employ \gptfour to compare translations generated by different models and assess them using the \textit{\elo rating system} \citep{elo1967proposed}. To the best of our knowledge, our work is the first attempt of adopting the widely-used \elo rating system in LLM evaluation to machine translation research. 
%     % We publicly release the annotations given by \gptfour to facilitate future research.
% \end{itemize}

\section{Related Work}
\paragraph{Document-Level Machine Translation}
In recent years, numerous approaches have been proposed for document-level machine translation (\docnmt). 
% The widely used baseline approach consists of simply concatenating a few adjacent sentences and feeding this as an input to the MT system, without modifying the system architecture in any way \citep{tiedemann-scherrer-2017-neural,bawden-etal-2018-evaluating, nguyen-etal-2021-data, sun-etal-2022-rethinking}. Apart from the simple concatenation method, 
There exist other approaches to \docnmt, including document embedding \citep{mace-servan-2019-using, huo-etal-2020-diving}, multiple encoders \citep{wang-etal-2017-exploiting-cross, bawden-etal-2018-evaluating, voita-etal-2018-context, zhang-etal-2018-improving}, attention variations \citep{miculicich-etal-2018-document, zhang-etal-2020-long, maruf-etal-2019-selective, wong-etal-2020-contextual, wu-etal-2023-document}, and translation caches \citep{maruf-haffari-2018-document, tu-etal-2018-learning, feng-etal-2022-learn}. Furthermore, \citet{DBLP:journals/csur/MarufSH21} present a comprehensive survey of \docnmt.

\paragraph{Large Language Models}
Large language models (LLMs) have demonstrated remarkable proficiency across a wide range of Natural Language Processing (NLP) tasks \citep{NEURIPS2020_1457c0d6, DBLP:journals/corr/abs-2204-02311, DBLP:journals/corr/abs-2211-05100, DBLP:journals/corr/abs-2305-10403, DBLP:journals/corr/abs-2302-13971, DBLP:journals/corr/abs-2307-09288}. Furthermore, recent research has shown that supervised fine-tuning (SFT) and Reinforcement Learning from Human Feedback (RLHF) can significantly enhance their performance when following general language instructions \cite{mishra-etal-2022-cross, wang-etal-2022-super, 
% DBLP:conf/iclr/WeiBZGYLDDL22, DBLP:conf/nips/Ouyang0JAWMZASR22, DBLP:journals/corr/abs-2210-11416, DBLP:journals/corr/abs-2211-01786,DBLP:journals/corr/abs-2304-14402,
DBLP:journals/corr/abs-2305-14705, DBLP:journals/corr/abs-2305-15011, DBLP:journals/corr/abs-2307-03025,wu-etal-2024-lamini}. More recently, there is a growing body of work exploring the translation capabilities of LLMs \citep{DBLP:journals/corr/abs-2305-06575, DBLP:journals/corr/abs-2306-10968, DBLP:journals/corr/abs-2309-11674, robinson-etal-2023-chatgpt,wu2024perhaps}. However, it is important to note that these efforts have primarily focused on sentence-level machine translation (\sennmt) and have not delved into document-level machine translation (\docnmt). A noteworthy study in \docnmt is conducted by \citet{wang-etal-2023-document-level}, where they investigate the document-level translation capabilities of \chatgpt, making it the most closely related work to our work.

\paragraph{Ours}
In contrast to the work of \citet{wang-etal-2023-document-level}, who primarily investigate the use of \chatgpt for \docnmt through prompting techniques, our study concentrates on analyzing the effectiveness of parameter-efficient fine-tuning (PEFT) and full fine-tuning (FFT) methods on moderately-sized LLMs in the context of \docnmt.
% In contrast to the work of \citet{wang-etal-2023-document-level}, who primarily explore the translation capabilities of \chatgpt using prompting techniques, our research focuses on examining parameter-efficient fine-tuning (PEFT) and fully fine-tuning (FFT) on moderately-sized LLMs within the context of \docnmt. 

\section{Experimental Setup}
In this section, we detail our experimental setup, covering the training strategy (\autoref{sec:training}), datasets (\autoref{sec:dataset}), models (\autoref{sec:models}), and evaluation (\autoref{sec:eval}).

\subsection{Two-Stage Training}
\label{sec:training}
% \docnmt approaches typically begin by pre-training the translation model on sentence-level parallel corpora, subsequently refining it through fine-tuning on document-level parallel corpora \citep{voita-etal-2019-good, maruf-etal-2019-selective, ma-etal-2020-simple, sun-etal-2022-rethinking, wu-etal-2023-document}. More recently, \citet{DBLP:journals/corr/abs-2309-11674} propose a two-stage training strategy, which initially involves fine-tuning a LLM on monolingual text, followed by a second fine-tuning phase on parallel text. Given that most state-of-the-art open-sourced LLMs are trained on English-centric corpora, our approach begins with the fine-tuning of a LLM on monolingual documents, followed by fine-tuning on parallel documents. Following \citet{DBLP:journals/corr/abs-2309-11674}, we omit the step of fine-tuning on sentence-level parallel datasets. 

\docnmt approaches typically start by pre-training the translation model on parallel sentences and then fine-tuning it on parallel documents \citep{voita-etal-2019-good, ma-etal-2020-simple, wu-etal-2023-document}. Recently, \citet{DBLP:journals/corr/abs-2309-11674} proposed a two-stage training strategy. This involves initially fine-tuning a LLM on monolingual text, followed by a second fine-tuning phase on parallel text. Following \citet{DBLP:journals/corr/abs-2309-11674}, our approach begins with fine-tuning an LLM on monolingual documents, followed by fine-tuning on parallel documents. 

\paragraph{Fine-tuning on Monolingual Documents}
Existing LLMs are typically pre-trained on English-centric corpora. Recent research highlights that these LLMs often exhibit sub-optimal performance on multilingual benchmarks \citep{DBLP:journals/corr/abs-2305-15011, DBLP:journals/corr/abs-2309-08958, DBLP:journals/corr/abs-2211-05100}. To address this limitation, our initial step involves fine-tuning all the parameters of LLMs using monolingual data from the target languages.
% To address this limitation, our initial step involves fine-tuning LLMs using monolingual data from the languages relevant to the translation tasks, thereby enhancing their proficiency in these specific languages. 
% Note that we fine-tune all the model parameters with the sequence length of $4096$ tokens in this stage.

\paragraph{Fine-tuning on Parallel Documents}
We fine-tune the model on document-level parallel corpora in this stage. Following \citet{wang-etal-2023-survey}, we condition each sentence pair on its context, consisting of the three preceding consecutive sentence pairs. As demonstrated by \citet{wang-etal-2023-document-level}, the prompting strategy plays a significant role in translating documents using LLMs. However, they only investigate how the prompting strategies affect \chatgpt and \gptfour at the inference stage. In our study, we first delve into how these prompting strategies impact the fine-tuning process, as shown in \autoref{fig:prompt_types_lite}, and we present our findings in \autoref{sec:prompting}.

\subsection{Datasets}
\label{sec:dataset}

\paragraph{Parallel Documents}
Following \citet{zhang-etal-2022-multilingual}, we conduct experiments on IWSLT2017 translation tasks \citep{cettolo-etal-2017-overview}. IWSLT2017 comprises translation datasets sourced from TED talks, encompassing translations between English and nine other languages, including Arabic, German, French, Italian, Japanese, Korean, Dutch, Romanian, and Chinese. There are approximately $1.9K$ sentence-aligned parallel documents with about $240K$ sentences for each language pair. The dataset statistics can be found in \autoref{sec:data_stat}.

\paragraph{Monolingual Documents}
We gather monolingual documents for all the target languages in our translation tasks, totaling ten languages. To manage computational limitations and address concerns about catastrophic forgetting that might result from excessive continued training, we leverage the data pruning technique suggested by \citet{DBLP:journals/corr/abs-2309-04564} to select $100M$ tokens for each language, including English, from the CulturaX corpus \citep{DBLP:journals/corr/abs-2309-09400}, totaling $1B$ tokens.

% We gather monolingual documents for all languages relevant to our downstream translation tasks, totaling ten languages. To manage computational limitations and address concerns about catastrophic forgetting that might result from excessive continued training, we limit each language to a collection of $100M$ tokens, resulting in a combined total of $1B$ tokens in monolingual documents. We le

% Inspired by \citet{DBLP:journals/corr/abs-2309-04564} that prunes the pre-training corpus with the perplexity given by LLMs, our process begins by randomly sampling $1B$ tokens for each language from the CulturaX dataset \citep{DBLP:journals/corr/abs-2309-09400}, and then we prune the sampled dataset using perplexity scores generated by the \model{Llama2-7B}. We only sample the documents with more than $2048$ tokens to maintain document coherence, and compute the perplexity of the first $1024$ tokens to reduce the computational costs. We then rank the sampled documents according to their perplexity scores, retaining only those falling within the $45^{\textrm{th}}$ to $55^{\textrm{th}}$ percentile range.

\begin{figure*}[t]
    \centering
    \begin{subfigure}[t]{0.47\textwidth}
    \centering
    \tiny
    \begin{Verbatim}[frame=single, fontsize=\tiny, breaklines=true, breakanywhere=true, commandchars=\\\{\}]
[<src_lang> Context]: <src1> <src2> <src3>
[<tgt_lang> Context]: <tgt1> <tgt2> <tgt3>
[<src_lang> Sentence]: <src4>
[<tgt_lang> Sentence]: <tgt4>
    \end{Verbatim}
    \caption{Prompt 1}
    \label{fig:prompt_one_lite}
    \end{subfigure}%
    \hfill
    \begin{subfigure}[t]{0.47\textwidth}
    \centering
    \tiny
    \begin{Verbatim}[frame=single, fontsize=\tiny, breaklines=true, breakanywhere=true]
[<src_lang>]: <src1> [<tgt_lang>]: <tgt1>
[<src_lang>]: <src2> [<tgt_lang>]: <tgt2>
[<src_lang>]: <src3> [<tgt_lang>]: <tgt3>
[<src_lang>]: <src4> [<tgt_lang>]: <tgt4>
    \end{Verbatim}
    \caption{Prompt 2}
    \label{fig:prompt_two_lite}
    \end{subfigure}%
    \vfill
    \begin{subfigure}[t]{0.47\textwidth}
    \centering
    \tiny
    \begin{Verbatim}[frame=single, fontsize=\tiny, breaklines=true, breakanywhere=true, commandchars=\\\{\}]
[<src_lang> Context]: <src1> <src2> <src3>
[<tgt_lang> Context]: <tgt1> <tgt2> <tgt3>
\textcolor{black}{Given the provided parallel context, translate the following <src_lang> sentence to <tgt_lang>:}
[<src_lang> Sentence]: <src4>
[<tgt_lang> Sentence]: <tgt4>
    \end{Verbatim}
    \caption{Prompt 3}
    \label{fig:prompt_three_lite}
    \end{subfigure}%
    \hfill
    \begin{subfigure}[t]{0.47\textwidth}
    \centering
    \tiny
    \begin{Verbatim}[frame=single, fontsize=\tiny, breaklines=true, breakanywhere=true, commandchars=\\\{\}]
[<src_lang>]: <src1> [<tgt_lang>]: <tgt1>
[<src_lang>]: <src2> [<tgt_lang>]: <tgt2>
[<src_lang>]: <src3> [<tgt_lang>]: <tgt3>
\textcolor{black}{Given the provided parallel sentence pairs, translate the following <src_lang> sentence to <tgt_lang>:}
[<src_lang>]: <src4> [<tgt_lang>]: <tgt4>
    \end{Verbatim}
    \caption{Prompt 4}
    \label{fig:prompt_four_lite}
    \end{subfigure}%

    \caption{
    Prompt types used in the preliminary study. 
    \texttt{<src\_lang>} and \texttt{<tgt\_lang>} indicate the source and target languages. \texttt{<src*>} and \texttt{<tgt*>} indicate the source and target sentences.
    \textbf{Note that the target sentences \texttt{<tgt*>} are only used during training and are replaced with the hypotheses \texttt{<hyp*>} generated by the model during inference.}
    Concrete examples for each prompt variation can be found in \autoref{sec:prompt_types}.
    }
    \label{fig:prompt_types_lite}
\end{figure*}

\subsection{Models}
\label{sec:models}

\paragraph{Baselines}
The baseline models in this study can be classified into three categories, including state-of-the-art LLMs and \sennmt models, and our re-implemented \docnmt models:
\begin{itemize}
    \item \textbf{State-of-the-art \sennmt models}: Our selection includes models such as \nllb, which are available with three different sets of parameters: 600M, 1.3B, and 3.3B.\footnote{Model signatures: \texttt{facebook/nllb-200-distilled-600M}, \texttt{facebook/nllb-200-1.3B}, and \texttt{facebook/nllb-200-3.3B}.} We also incorporate the widely-used commercial translation system, Google Translate.
    \item \textbf{State-of-the-art LLMs}: For our baseline LLMs in the context of \docnmt, we utilize \chatgpt and \gptfour.\footnote{Model signatures: \texttt{gpt-3.5-turbo-1106} and \texttt{gpt-4-1106-preview}.} We use the Prompt 4 as detailed in \autoref{fig:prompt_four_lite} during the translation process.
    \item \textbf{Our re-implemented \docnmt models}: We conduct full fine-tuning on the concatenation-based \docnmt model \citep{tiedemann-scherrer-2017-neural}, as well as several recent \docnmt baselines \citep{sun-etal-2022-rethinking, wu-etal-2023-document, wu-etal-2024-importance}, initialized with \mtfive \citep{xue-etal-2021-mt5}. These models are available with parameters of 300M, 580M, and 1.2B, representing the strong \docnmt baseline.
\end{itemize}

\paragraph{Ours}
In this work, we utilize \model{Llama2-7B}, \model{Bloom-7B}, and \model{Vicuna-7B}, as our backbones.\footnote{\model{Llama2} signature: \texttt{meta-llama/Llama-2-7b-hf}, \model{Bloom} signature: \texttt{bigscience/bloom-7b1}, and \model{Vicuna} signature: \texttt{lmsys/vicuna-7b-v1.5}. Note that \model{Vicuna}-v1.5 models are fine-tuned from \model{Llama2}.} The \model{Llama2} models are predominantly pre-trained on English text, while the \model{Bloom} models are pre-trained on multilingual text. The use of \model{Vicuna} models allows us to compare the differences between base models and instruction-tuned models (\model{Llama2} vs. \model{Vicuna}). 
% We firstly fine-tune all the model parameters of these models on the collected monolingual documents, and then fine-tune these models on the parallel documents. 
We denote those fully fine-tuned models as \oursllamasmallfft, \oursbloomsmallfft, and \oursvicunasmallfft. We denote those models fine-tuned with \model{LoRA} \citep{DBLP:conf/iclr/HuSWALWWC22} as \oursllamasmalllora, \oursbloomsmalllora, and \oursvicunasmalllora. 
The optimization details can be found in \autoref{sec:optim}.

\subsection{Evaluation}
\label{sec:eval}

\paragraph{Evaluation Metrics}
We evaluate the translation on the IWSLT2017 test sets \citep{cettolo-etal-2017-overview} using sentence-level \model{BLEU} \citep{papineni-etal-2002-bleu} and document-level \model{BLEU} \citep{liu-etal-2020-multilingual-denoising} using SacreBLEU \citep{post-2018-call}, denoted as \sbleu and \dbleu.\footnote{BLEU signature: \texttt{nrefs:1|case:mixed|eff:no|\\tok:[13a|ja-mecab-0.996-IPA|ko-mecab-0.996/ko\\-0.9.2-KO|zh]|smooth:exp|version:2.3.1}.} Furthermore, as conventional MT metrics like BLEU demonstrate poor correlation to human judgments \citep{freitag-etal-2022-results}, we also evaluate the translation quality with the state-of-the-art neural evaluation metric \comet \cite{rei-etal-2020-comet}.\footnote{COMET signature: \texttt{Unbabel/wmt22-comet-da}.} Moreover, we use the average sentence-level \model{BLEU} \avgsbleu, the average document-level \model{BLEU} \avgdbleu, and the average \comet \avgcomet for the overall performance.

\paragraph{Settings}
We evaluate our models in two settings: \textit{Bilingual English-from/to-Many Translations} and \textit{Multilingual English-from/to-Many Translations}. For the bilingual translations, we train our models on the training set for one specific language pair. For the multilingual translations, we combine all the available training sets for training.

\paragraph{Inference}
In this work, we explore two inference strategies. The first strategy involves using previous translations as context for the current translation, translating test examples in their original order. This begins with the first sentence, which is free from context, as illustrated in \autoref{fig:prompt_four_lite}. We refer to this strategy as \textbf{\model{ReUse}}. Alternatively, sentences within the context can be re-generated individually for each document translation to prevent error propagation during the translation process. We call this strategy \textbf{\model{ReGen}}. For reasons of inference efficiency, \model{ReUse} is used as the default inference strategy throughout this study, unless specified otherwise. A comparison of these two inference strategies is provided in \autoref{sec:analysis}.

\begin{table}[t]
\centering
\scriptsize
\setlength{\tabcolsep}{8pt}
\begin{tabular}{lcccc}
\toprule
                     & PID & \avgsbleu     & \avgdbleu     & \avgcomet     \\ \midrule
\oursllamasmalllora  & 1   & 15.5          & 18.2          & 67.5          \\
                     & 2   & 19.0          & 21.9          & 70.7          \\
                     & 3   & 15.8          & 18.3          & 69.8          \\
                     & 4   & \textbf{20.2} & \textbf{23.4} & \textbf{72.7} \\ \midrule
\oursbloomsmalllora  & 1   & 19.3          & 20.5          & 70.5          \\
                     & 2   & 20.6          & 23.5          & 73.6          \\
                     & 3   & 19.8          & 20.8          & 73.9          \\
                     & 4   & \textbf{23.1} & \textbf{27.3} & \textbf{76.8} \\ \midrule
\oursvicunasmalllora & 1   & 19.0          & 22.4          & 74.2          \\
                     & 2   & 20.4          & 23.5          & 71.6          \\
                     & 3   & 18.3          & 21.4          & 70.0          \\
                     & 4   & \textbf{22.4} & \textbf{25.7} & \textbf{76.2} \\ \bottomrule
\end{tabular}
\caption{
Overall performance given by \oursllamasmalllora, \oursbloomsmalllora, and \oursvicunasmalllora on different prompt variations, across four English-centric translation tasks involving German and Chinese.
PID indicates the prompt ID in \autoref{fig:prompt_types_lite}.
Best results are highlighted in \textbf{bold}.
}
\label{tab:prompt_types_overall}
\end{table}

\section{A Preliminary Study on Prompts}
\label{sec:prompting}

% Prompt is of vital importance in the research of LLMs. Recent studies demonstrate that an ideal prompt can significantly boost the model performance and unlock surprising model capabilities \citep{DBLP:conf/nips/KojimaGRMI22, DBLP:conf/nips/Wei0SBIXCLZ22}. Therefore, we firstly investigate impact of the prompt in the process of fine-tuning.
The prompt plays a crucial role in LLM research. Recent studies show that an optimal prompt can greatly enhance model performance and reveal unexpected model capabilities \citep{DBLP:conf/nips/KojimaGRMI22, DBLP:conf/nips/Wei0SBIXCLZ22}. Hence, our initial focus is on investigating the prompt's impact during fine-tuning.

\paragraph{Prompt Variations}
% As shown in \autoref{fig:prompt_types_lite}, we design four types of prompts in our preliminary study. 
% With different designs of prompts, we aim to address two research questions: \textit{How does the context structure affect the translation quality?} (Prompt 1 vs. Prompt 2) and \textit{How do the natural language instructions affect the translation quality?} (Prompt 1 vs. Prompt 3). Furthermore, we explore the compound effect of these two aspects of our prompt designs in Prompt 4 (\autoref{fig:prompt_four_lite}).

Displayed in \autoref{fig:prompt_types_lite}, our preliminary study features four prompt types. These designs aim to tackle two research questions: \textit{How does context structure impact translation quality?} (Prompt 1 vs. Prompt 2) and \textit{How do natural language instructions influence translation quality?} (Prompt 1 vs. Prompt 3). We also investigate the combined effect of these aspects in Prompt 4.

\begin{table*}[t]
\centering
\small
\setlength{\tabcolsep}{5pt}
\begin{tabular}{lcccccccc}
\toprule
                     & \multirow{2}{*}{\# of param.} & \# of train.    & \multicolumn{3}{c}{En-X}                                                          & \multicolumn{3}{c}{X-En}                                                            \\ \cmidrule(rl){4-6} \cmidrule(l){7-9}
                     &                               & param.          & \avgsbleu                 & \avgdbleu                 & \avgcomet                 & \avgsbleu                  & \avgdbleu                  & \avgcomet                 \\ \midrule
\multicolumn{9}{l}{\cellcolor{gray!30}\textit{State-of-the-art \sennmt baselines}}                                                                                                                                                                                  \\
\nllb                & 600M                          & ---             & 23.6                      & 27.3                      & 82.3                      & 18.2                       & 22.1                       & 72.8                      \\
                     & \phantom{0}1.3B               & ---             & 25.7                      & 29.5                      & 83.5                      & 25.0                       & 28.7                       & 78.1                      \\
                     & \phantom{0}3.3B               & ---             & \underline{26.8}          & \underline{30.5}          & \underline{84.3}          & \underline{25.8}           & \underline{29.4}           & 78.9                      \\
\googletrans         & ---                           & ---             & 24.5                      & 28.4                      & 81.6                      & 25.0                       & 28.5                       & \underline{81.2}          \\ \midrule
\multicolumn{9}{l}{\cellcolor{gray!30}\textit{State-of-the-art LLMs}}                                                                                                                                                                                               \\
\chatgpt             & ---                           & ---             & 26.3                      & 30.1                      & 85.3                      & 30.7                       & 34.1                       & 85.5                      \\
\gptfour             & ---                           & ---             & \textbf{\underline{27.0}} & \textbf{\underline{30.7}} & \textbf{\underline{86.3}} & \textbf{\underline{31.7}}  & \textbf{\underline{35.1}}  & \textbf{\underline{86.0}} \\ \midrule
\multicolumn{9}{l}{\cellcolor{gray!30}\textit{LLM backbones}}                                                                                                                                                                                                       \\
\model{Llama2-7B}    & 7B                           & ---             & \phantom{0}2.7            & \phantom{0}3.5            & 40.1                      & \phantom{0}4.2             & \phantom{0}4.4             & 52.2                      \\
\model{Bloom-7B}     & 7B                           & ---             & \phantom{0}2.5            & \phantom{0}2.9            & 35.5                      & \phantom{0}6.7             & \phantom{0}7.3             & 49.4                      \\
\model{Vicuna-7B}    & 7B                           & ---             & \underline{10.2}          & \underline{12.4}          & \underline{64.7}          & \phantom{0}\underline{9.5} & \phantom{0}\underline{9.8} & \underline{62.9}          \\ \midrule
\multicolumn{9}{l}{\cellcolor{gray!30}\textit{Re-implemented \docnmt baselines}}                                                                                                                                                                                    \\
\docnmtmtfive \citeyearpar{tiedemann-scherrer-2017-neural}        & 300M                          & 300M            & 17.2                      & 20.2                      & 75.1                      & 19.4                       & 21.2                       & 75.1                      \\
                     & 580M                          & 580M            & 18.6                      & 21.5                      & 78.3                      & 20.7                       & 22.5                       & 77.4                      \\
                     & \phantom{0}1.2B               & \phantom{0}1.2B & 18.4                      & 21.4                      & 79.2                      & 21.5                       & 23.4                       & 78.7                      \\
\mrdoctosent \citeyearpar{sun-etal-2022-rethinking}         & \phantom{0}1.2B               & \phantom{0}1.2B & 18.8                      & 21.9                      & 79.9                      & 22.0                       & 23.8                       & 79.3                      \\
\mrdoctodoc \citeyearpar{sun-etal-2022-rethinking}          & \phantom{0}1.2B               & \phantom{0}1.2B & ---                       & \underline{22.5}          & ---                       & ---                        & 24.0                       & ---               \\
\docflat \citeyearpar{wu-etal-2023-document}             & \phantom{0}1.2B               & \phantom{0}1.2B & 19.2                      & 22.4                      & 80.2                      & \underline{22.2}           & \underline{24.3}           & 79.3                      \\
\iada \citeyearpar{wu2024importance}               & \phantom{0}1.2B               & \phantom{0}1.2B & \underline{19.3}          & 22.4                      & \underline{80.4}          & 22.1                       & 24.0                       & \underline{79.5}                      \\ \midrule
\multicolumn{9}{l}{\cellcolor{gray!30}\textit{Bilingual English-from/to-Many LLM-based \docnmt models (Ours)}}                                                                                                                                                                                     \\
\oursllamasmalllora  & \phantom{00.}7B               & \phantom{00}8M  & 17.2                      & 20.2                      & \underline{70.8}          & 23.8                       & 25.7                       & 73.7                      \\
\oursllamasmallfft   & \phantom{00.}7B               & \phantom{00.}7B & 13.7                      & 16.2                      & 67.4                      & 22.4                       & 24.1                       & 74.0                      \\
\oursbloomsmalllora  & \phantom{00.}7B               & \phantom{00}8M  & \underline{17.7}          & \underline{20.5}          & 68.5                      & \underline{29.9}           & \underline{33.6}           & \underline{81.4}          \\
\oursbloomsmallfft   & \phantom{00.}7B               & \phantom{00.}7B & 12.0                      & 13.8                      & 59.6                      & 22.3                       & 24.5                       & 69.9                      \\
\oursvicunasmalllora & \phantom{00.}7B               & \phantom{00}8M  & 15.8                      & 18.6                      & 68.8                      & 21.6                       & 23.3                       & 71.4                      \\
\oursvicunasmallfft  & \phantom{00.}7B               & \phantom{00.}7B & 14.3                      & 16.8                      & 65.0                      & 21.8                       & 23.5                       & 74.3                      \\ \midrule
\multicolumn{9}{l}{\cellcolor{gray!30}\textit{Multilingual English-from/to-Many LLM-based \docnmt models (Ours)}}                                                                                                                                                                                     \\
\oursllamasmalllora  & \phantom{00.}7B               & \phantom{00}8M  & 13.9                      & 16.9                      & 67.2          & 17.1                       & 18.4                       & 62.0                      \\
\oursllamasmallfft   & \phantom{00.}7B               & \phantom{00.}7B & \underline{17.1}                      & \underline{20.4}                      & \underline{73.8}                      & 18.3                       & 19.5                       & 69.6                      \\
\oursbloomsmalllora  & \phantom{00.}7B               & \phantom{00}8M  & 10.3          & 12.3          & 58.1                      & 23.5           & 25.7           & 75.8          \\
\oursbloomsmallfft   & \phantom{00.}7B               & \phantom{00.}7B & 16.0                      & 18.6                      & 67.8                      & \underline{27.2}                       & \underline{30.5}                       & \underline{78.7}                      \\
\oursvicunasmalllora & \phantom{00.}7B               & \phantom{00}8M  & 12.9                      & 15.7                      & 66.4                      & 17.2                       & 18.6                       & 60.6                      \\
\oursvicunasmallfft  & \phantom{00.}7B               & \phantom{00.}7B & 13.8                      & 16.8                      & 69.3                      & 18.4                       & 19.4                       & 65.6                      \\ \bottomrule
\end{tabular}
\caption{
Overall performance on IWSLT2017. 
\# of param. indicates the number of parameters of the model.
\# of train. param. indicates the number of trainable parameters of the model.
All the LLM approaches use Prompt 4 (\autoref{fig:prompt_four_lite}) during inference.
Best results are highlighted in \textbf{bold}.
Best results in each group are \underline{underlined}.
% \trang{Perhaps highlight the best results in each block with underscore. It would make the main results section easier to follow.}
}
\label{tab:main}
\end{table*}

\paragraph{Results}

Our investigation analyzes prompt variations using three PEFT models (\oursllamasmalllora, \oursbloomsmalllora, and \oursvicunasmalllora) on four English-centric translation tasks involving German and Chinese. Overall results are presented in \autoref{tab:prompt_types_overall}. Comparing Prompt 1 (\autoref{fig:prompt_one_lite}) and Prompt 2 (\autoref{fig:prompt_two_lite}), we find that models fine-tuned with Prompt 2 generally outperform those with Prompt 1, indicating Prompt 2's effectiveness in enhancing LLM performance. Regarding our second research question (\autoref{fig:prompt_one_lite} vs. \autoref{fig:prompt_three_lite}), we observe varied performance. \oursllamasmalllora and \oursbloomsmalllora perform better with Prompt 3, while \oursvicunasmalllora performs better with Prompt 1. These results highlight varying impacts of prompt variations across models and suggest natural language instructions are less effective when using instruction-tuned language models as model backbones. Finally, LLMs with Prompt 4 (\autoref{fig:prompt_four_lite}) achieve the best overall performance, suggesting a positive compound effect of context structure and instructions.

\paragraph{Conclusion}
As expected, the prompt plays a significant role in LLM performance. A well-structured prompt, combining appropriate context and clear instructions, can significantly boost model performance. In this work, we use Prompt 4 (\autoref{fig:prompt_four_lite}) in our experiments.

\begin{figure*}[t]
    \centering
    \begin{subfigure}[t]{0.47\textwidth}
    \centering
    \includegraphics[scale=0.3]{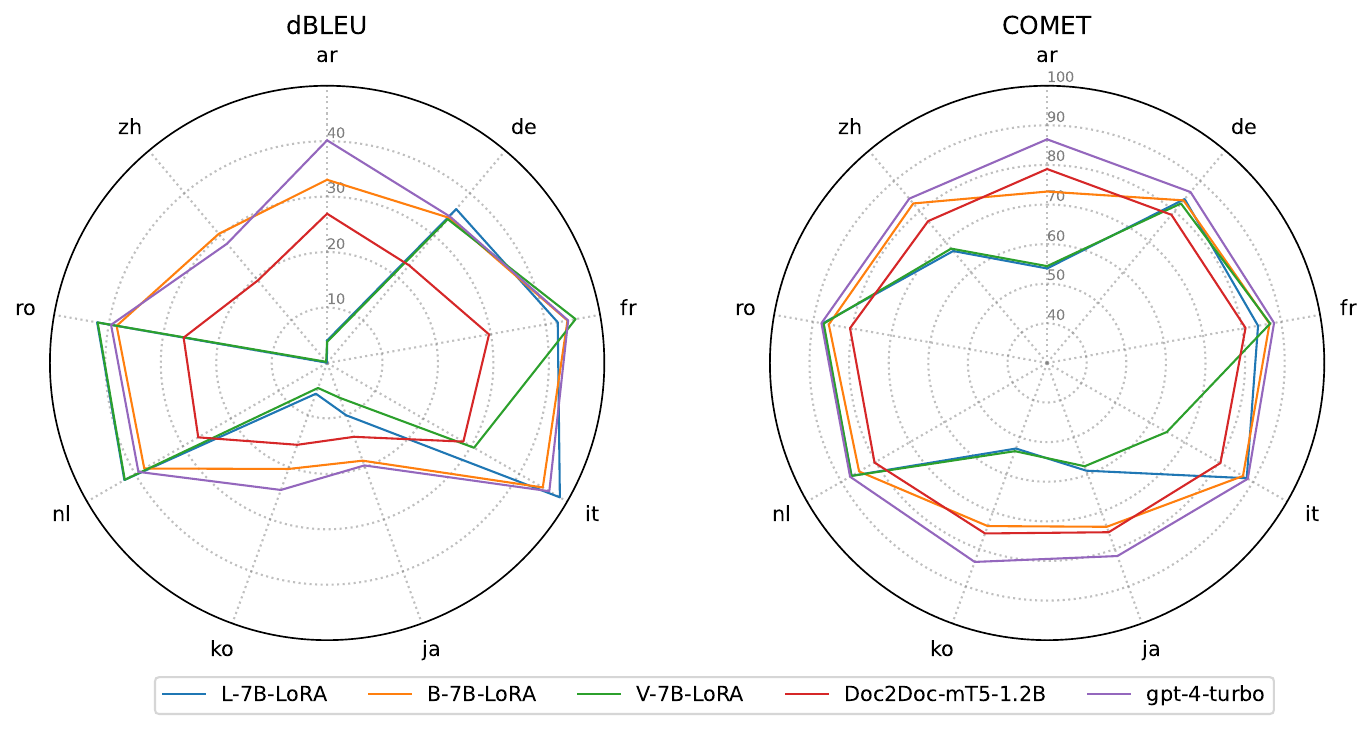}
    \caption{
    Bilingual English-from-Many translations.
    }
    \label{fig:radars_bilingual_x_en}
    \end{subfigure}%
    \hfill
    \begin{subfigure}[t]{0.47\textwidth}
    \centering
    \includegraphics[scale=0.3]{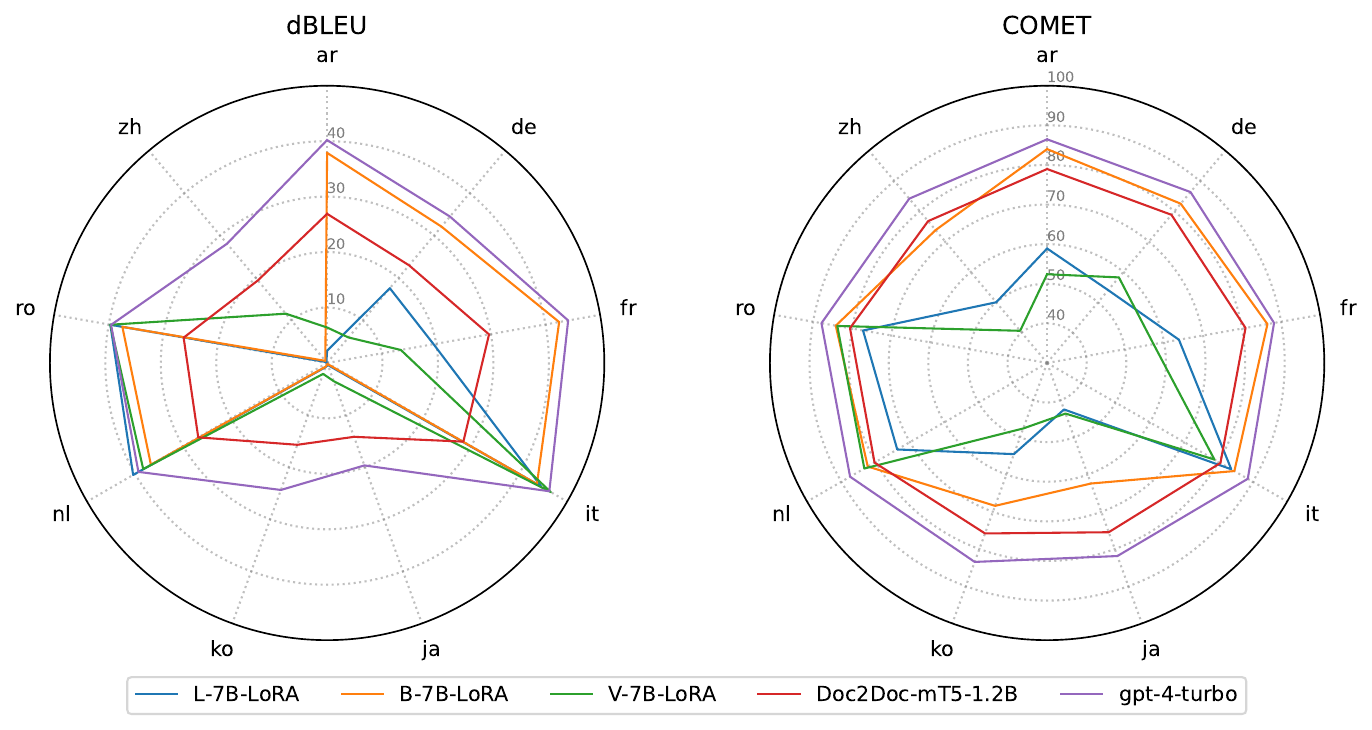}
    \caption{
    Multilingual English-from-Many translations.
    }
    \label{fig:radars_multilingual_x_en}
    \end{subfigure}%

    \caption{
    Breakdown results on \sbleu, \dbleu, and \comet given by \oursllamasmalllora, \oursvicunasmalllora, \oursbloomsmalllora, \docnmtmtfivelarge, and \gptfour for the bilingual (\autoref{fig:radars_bilingual_x_en}) and multilingual (\autoref{fig:radars_multilingual_x_en}) translation tasks \textbf{from other languages to English}.
    }
    \label{fig:radars_x_en}
\end{figure*}
\section{Main Results}
\label{sec:main_results}

\paragraph{Overall Performance}
% In our results presented in \autoref{tab:main}, we observe that \gptfour and \chatgpt significantly outshine all other models in performance. Notably, the \nllb variants, which are trained on vast amount of parallel sentence pairs, also demonstrate superior performance among specialized machine translation (MT) models. In the context of \docnmt, conventional \docnmt models still outperform our LLM-based \docnmt models for translations from English to other languages when evaluated using standard MT metrics. Conversely, for translations from other languages to English, our LLM-based \docnmt models perform on par or better than conventional \docnmt models in \avgsbleu and \avgdbleu metrics, while those conventional \docnmt models maintain superior performance in \avgcomet.

In our results presented in \autoref{tab:main}, we observe that \gptfour and \chatgpt significantly outperform all other models. Notably, the \nllb variants also demonstrate superior performance among specialized machine translation (MT) models. In the context of \docnmt, conventional \docnmt models still outperform our LLM-based \docnmt models for translations from English to other languages when evaluated using standard MT metrics. Conversely, for translations from other languages to English, our LLM-based \docnmt models perform on par or better than conventional \docnmt models in \avgsbleu and \avgdbleu metrics, while conventional \docnmt models maintain superior performance in \avgcomet.

\paragraph{LLM-based \docnmt Models}
As shown in \autoref{tab:main}, models using \model{LoRA} generally outperform fully fine-tuned (FFT) LLMs in bilingual translations, while fully fine-tuned LLMs surpass \model{LoRA} models in multilingual translations. The \model{LoRA} approach, which updates only a limited number of parameters, allows the model to retain the rich linguistic knowledge of the pre-trained LLM while adapting its capabilities to a particular bilingual context. However, in multilingual translation tasks, this parameter-efficient fine-tuning is insufficient to capture the complexities of multiple languages simultaneously. Conversely, fully fine-tuning all the parameters of LLMs enables better capture of multilingual complexity but makes the model prone to overfitting in bilingual translations.

% This discrepancy is likely due to \textit{catastrophic forgetting}. During extensive fine-tuning with a large volume of parallel documents, full fine-tuning tends to erode much of the pre-trained LLM's knowledge and capabilities.

% We present the results for the translation tasks from other languages to English in \autoref{fig:radars_x_en}. Regarding the readability of the figures, we present only the results provided by our models using \model{LoRA}. Our LLM-based \docnmt models exhibit superior performance, sometimes even surpassing \gptfour in certain translation tasks. However, they fail completely in others. A manual review of translation tasks where our LLM-based \docnmt models fail reveals that the primary cause of failure is \textit{off-target translation}. We provide an in-depth analysis of the off-target translation problem in \autoref{sec:analysis}. A complete breakdown of the results is in \autoref{sec:breakdown}.

\paragraph{Breakdown Performance}
We present the results for translation tasks from other languages to English in \autoref{fig:radars_x_en}. For clarity, only the results from our models using \model{LoRA} are shown. We observe that LLM-based \docnmt models sometimes even surpassing \gptfour in certain bilingual and multilingual translation tasks, though they fail completely in others. Our manual review indicates that the primary cause of these failures is \textit{off-target translation}. An in-depth analysis of this issue is provided in \autoref{sec:analysis}, and a complete breakdown of the results is available in \autoref{sec:breakdown}.

\section{Analysis}
\label{sec:analysis}
In this section, we examine off-target issues with decoding strategy and use GPT-4 to analyze translation errors. We also explore discourse phenomena, training strategies, the data efficiency of parallel documents, and additional evaluations on recent WMT2023 test sets. Additionally, we evaluate the zero-shot crosslingual transfer capabilities of our models and present the results in \autoref{sec:transfer}.

\begin{table}[t]
\centering
\small
\setlength{\tabcolsep}{7pt}
\begin{tabular}{lccccc}
\toprule
                     & $\mu_{\%}$     & Ar             & Ja           & Ko             & Zh             \\ \midrule
\oursllamasmalllora  & 29.2           & 87.9           & 25.5         & 44.2           & 93.1           \\
\oursllamasmallfft   & 40.2           & 87.9           & 75.5         & 92.3           & 93.6           \\
\oursbloomsmalllora  & \phantom{0}2.8 & \phantom{0}2.9 & \phantom{0}4.0 & \phantom{0}8.4 & \phantom{0}1.6 \\
\oursbloomsmallfft   & 28.0           & 54.1           & 43.8         & 70.4           & 76.4           \\
\oursvicunasmalllora & 32.3           & 88.2           & 40.4         & 35.7           & 90.5           \\
\oursvicunasmallfft  & 44.7           & 94.1           & 98.3         & 96.6           & 94.6           \\ \bottomrule
\end{tabular}
\caption{
Off-target rate (\%) provided by our LLM-based \docnmt models for translation tasks from selective languages to English.
$\mu_{\%}$ indicates the average off-target rate across all nine language pairs.
\textbf{A lower off-target rate indicates better performance}.
The complete results are presented in \autoref{sec:off_target_app}.
}
\label{tab:off_target_x_en_lite}
\end{table}

% \label{sec:off_target}
\paragraph{Off-Target Translation}
In \autoref{fig:radars_x_en}, our LLM-based \docnmt models excel in some translation tasks but struggle in others due to off-target translation issues. We investigate this problem using the \textit{fasttext} library \citep{bojanowski-etal-2017-enriching} to identify translation languages and quantify off-target rates, which represent the proportion of translations that are off-target. Results are presented in \autoref{tab:off_target_x_en_lite}, with off-target rates reaching up to 98.3\% in failing tasks. Notably, only \oursbloomsmalllora consistently maintains low off-target rates, likely due to \model{Bloom-7B}'s multilingual pre-training. Detailed off-target rates are provided in \autoref{sec:off_target_app}.

\begin{table}[t]
\centering
\small
\setlength{\tabcolsep}{6pt}
\begin{tabular}{lccc}
\toprule
                                    & \sbleu         & \dbleu         & \comet \\ \midrule
\oursllamasmalllora + \model{ReUse} & \phantom{0}3.9 & \phantom{0}4.1 & 53.9   \\
\oursllamasmalllora + \model{ReGen} & 17.5           & 19.6           & 69.7   \\ \midrule
\oursllamasmallfft + \model{ReUse}  & \phantom{0}2.5 & \phantom{0}2.6 & 51.6   \\
\oursllamasmallfft + \model{ReGen}  & 15.9           & 17.4           & 65.0   \\ \bottomrule
\end{tabular}
\caption{Results on Arabic-English translation given by different decoding strategies.}
\label{tab:decode}
\end{table}
\paragraph{Inference Strategy}
As described in \autoref{sec:eval}, previous translations provide context for the current translation. We hypothesize that the high off-target translation rate in \autoref{tab:off_target_x_en_lite} is due to error propagation during the decoding stage. Therefore, we refer to the decoding strategy in \autoref{sec:eval} as \model{ReUse}, and introduce an alternative strategy where all context translations are re-generated, called \model{ReGen}. As shown in \autoref{tab:decode}, the \model{ReGen} strategy significantly improves translation quality, confirming our hypothesis on decoding error propagation. However, the inference cost of \model{ReGen} is four times higher than that of \model{ReUse}. These findings highlight the main reason for translation failures in LLM-based \docnmt models.

\begin{figure}[t]
    \centering
    \begin{tikzpicture}[scale=0.7]
      \begin{axis}[
        width = 0.8\linewidth,
        height = 8.5cm,
        xbar,
        tickwidth         = 0pt,
        bar width = 2pt,
        enlarge y limits  = 0.05,
        enlarge x limits  = 0.05,
        tick label style={font=\small},
        every node near coord/.append style={font=\scriptsize},
        symbolic y coords = {
            \textcolor{orange}{Multiple terms in translation},
            \textcolor{orange}{Inconsistent style},
            \textcolor{orange}{Coherence},
            \textcolor{orange}{Cohesion},
            % \textcolor{orange}{Unclear reference},
            % Duplication,
            % Spelling,
            % Punctuation,
            Grammar,
            % Completeness,
            % Retained factual error,
            % Untranslated,
            % Do not translate,
            % Unjustfied euphemism,
            Omission,
            Addition,
            Undertranslation,
            Overtranslation,
            Mistranslation,
        },
        nodes near coords,
        xlabel={\# of occurences},
        ytick=data,
        xmin=0, xmax=2400,
        legend style={
            at={(0.5,-0.15)},
            anchor=north,
            legend columns=2,
        },
        legend cell align={left}, 
      ]

        % l-7b-lora
        \addplot[style={red,fill=red,mark=none}] coordinates {
        (1395,Mistranslation)
        (717,Overtranslation)
        (1224,Undertranslation)
        (840,Addition)
        (484,Omission)
        % (4,Unjustfied euphemism)
        % (39,Do not translate)
        % (52,Untranslated)
        % (5,Retained factual error)
        % (9,Completeness)
        (234,Grammar)
        % (66,Punctuation)
        % (24,Spelling)
        % (93,Duplication)
        % (64,\textcolor{orange}{Unclear reference})
        (709,\textcolor{orange}{Cohesion})
        (454,\textcolor{orange}{Coherence})
        (725,\textcolor{orange}{Inconsistent style})
        (346,\textcolor{orange}{Multiple terms in translation})
        };
        \addlegendentry{\oursllamasmalllora}

        % l-7b-fft
        \addplot[style={blue,fill=blue,mark=none}] coordinates {
        (1356,Mistranslation)
        (717,Overtranslation)
        (1358,Undertranslation)
        (863,Addition)
        (616,Omission)
        % (2,Unjustfied euphemism)
        % (17,Do not translate)
        % (45,Untranslated)
        % (3,Retained factual error)
        % (3,Completeness)
        (266,Grammar)
        % (44,Punctuation)
        % (35,Spelling)
        % (46,Duplication)
        % (102,\textcolor{orange}{Unclear reference})
        (727,\textcolor{orange}{Cohesion})
        (565,\textcolor{orange}{Coherence})
        (771,\textcolor{orange}{Inconsistent style})
        (339,\textcolor{orange}{Multiple terms in translation})
        };
        \addlegendentry{\oursllamasmallfft}

        % google translate
        \addplot[style={gray,fill=gray,mark=none}] coordinates {
        (1712,Mistranslation)
        (715,Overtranslation)
        (977,Undertranslation)
        (245,Addition)
        (417,Omission)
        % (2,Unjustfied euphemism)
        % (45,Do not translate)
        % (11,Untranslated)
        % (4,Retained factual error)
        % (5,Completeness)
        (486,Grammar)
        % (203,Punctuation)
        % (109,Spelling)
        % (105,Duplication)
        % (80,\textcolor{orange}{Unclear reference})
        (804,\textcolor{orange}{Cohesion})
        (592,\textcolor{orange}{Coherence})
        (824,\textcolor{orange}{Inconsistent style})
        (273,\textcolor{orange}{Multiple terms in translation})
        };
        \addlegendentry{\googletrans}

        % mt5 large
        \addplot[style={green,fill=green,mark=none}] coordinates {
        (2002,Mistranslation)
        (836,Overtranslation)
        (1551,Undertranslation)
        (586,Addition)
        (868,Omission)
        % (0,Unjustfied euphemism)
        % (25,Do not translate)
        % (22,Untranslated)
        % (10,Retained factual error)
        % (1,Completeness)
        (474,Grammar)
        % (80,Punctuation)
        % (128,Spelling)
        % (45,Duplication)
        % (206,\textcolor{orange}{Unclear reference})
        (933,\textcolor{orange}{Cohesion})
        (717,\textcolor{orange}{Coherence})
        (854,\textcolor{orange}{Inconsistent style})
        (386,\textcolor{orange}{Multiple terms in translation})
        };
        \addlegendentry{\docnmtmtfivelarge}

      % \addplot coordinates { (14320,Mistranslation)         (1615,Overtranslation)
      %                        (560,Distributions)   (3075,Editors)  };
      % \legend{\oursllamasmalllora, \oursllamasmallfft, \googletrans, \docnmtmtfivelarge}
      \end{axis}
    \end{tikzpicture}
    \caption{
    Error type analysis given by \gptfour for translations from English to German, Romanian, and Chinese. \textcolor{orange}{The error types in orange} are context-dependent.
    We omit those error types that are rare.
    }
    \label{fig:error_types}
\end{figure}
\paragraph{Translation Errors}
To understand the translation capabilities of our LLM-based \docnmt models, we select specific error types from the Multidimensional Quality Metrics (MQM) framework \citep{burchardt-2013-multidimensional}. \citet{kocmi-federmann-2023-gemba} demonstrate that \model{gpt-4} can identify error spans and achieve state-of-the-art MT evaluation accuracy, so we use \gptfour to analyze translation errors in texts translated by these models. Due to resource constraints, we focus on four models: \oursllamasmalllora, \oursllamasmallfft, \docnmtmtfivelarge, and \googletrans, assessing translations from English to German, Romanian, and Chinese. The error identification prompt is detailed in \autoref{sec:gpt_prompts}, and we present the frequency of error types in \autoref{fig:error_types}. Most errors are limited to individual sentences. Despite similar scores in metrics such as \sbleu, \dbleu, and \comet among the models, our LLM-based \docnmt models (\oursllamasmalllora and \oursllamasmallfft) exhibit fewer context-independent and context-dependent errors. This suggests that current evaluation metrics may not sufficiently assess document-level translations and indicates that fine-tuning LLMs holds promise for enhancing \docnmt performance.

\begin{table}[t]
\centering
\small
\setlength{\tabcolsep}{10pt}
\begin{tabular}{lcc}
\toprule
                     & En-De & En-Fr \\ \midrule
\docnmtmtfivelarge   & 77.0  & 89.9  \\
\oursllamasmalllora  & 83.1  & \textbf{95.1}  \\
\oursllamasmallfft   & 81.1  & 94.6  \\
\oursbloomsmalllora  & 75.5  & 91.9  \\
\oursbloomsmallfft   & 68.3  & 90.8  \\
\oursvicunasmalllora & \textbf{84.9}  & 94.8  \\
\oursvicunasmallfft  & 84.4  & 94.8  \\ \bottomrule
\end{tabular}
\caption{
Accuracy (in \%) on the English-German and English-French contrastive test sets.
Best results are highlighted in \textbf{bold}.
}
\label{tab:disco}
\end{table}
% \subsection{Discourse Phenomena}
\paragraph{Discourse Phenomena}
To evaluate our LLM-based \docnmt model's ability to leverage contextual information, we assessed it using the English-German contrastive test set by \citet{muller-etal-2018-large} and the English-French contrastive test set by \citet{lopes-etal-2020-document}. This evaluation tests the model's accuracy in selecting the correct pronoun from multiple translation options. Results, shown in \autoref{tab:disco}, reveal that models initialized with \model{Llama2-7B} and \model{Vicuna-7B} outperform \docnmtmtfivelarge, while \model{Bloom-7B}-initialized models perform worse, indicating that contextual understanding is mostly acquired during pre-training, as detailed by \citet{DBLP:journals/corr/abs-2211-05100} due to the lack of German text in \model{Bloom} pre-training. The \textit{generative accuracy} results following \citet{DBLP:journals/corr/abs-2304-12959} are presented in \autoref{sec:disco}.

\begin{table}[t]
\centering
\small
\setlength{\tabcolsep}{7pt}
\begin{tabular}{lccc}
\toprule
      & \sbleu         & \dbleu         & \comet \\ \midrule
\multicolumn{4}{l}{\cellcolor{gray!30}\textit{One-Stage}}           \\
Nl-En & 30.2           & 32.4           & 71.2   \\
Ro-En & 28.3           & 30.1           & 72.3   \\
Ar-En & \phantom{0}1.3 & \phantom{0}2.1 & 50.1   \\
Zh-En & \phantom{0}0.1 & \phantom{0}0.1 & 60.9   \\ \midrule
\multicolumn{4}{l}{\cellcolor{gray!30}\textit{Two-Stage}}           \\
Nl-En & 38.9           & 41.9           & 87.0   \\
Ro-En & 38.2           & 41.4           & 87.3   \\
Ar-En & \phantom{0}2.5 & \phantom{0}2.6 & 51.6   \\
Zh-En & \phantom{0}0.1 & \phantom{0}0.1 & 67.1   \\ \midrule
\multicolumn{4}{l}{\cellcolor{gray!30}\textit{Three-Stage}}         \\
Nl-En & 39.1           & 42.1           & 87.0   \\
Ro-En & 38.4           & 41.6           & 87.3   \\
Ar-En & \phantom{0}2.3 & \phantom{0}2.4 & 52.4   \\
Zh-En & \phantom{0}0.3 & \phantom{0}0.3 & 67.4   \\ \bottomrule
\end{tabular}
\caption{Results from one-stage, two-stage, and three-stage training strategies. The one-stage strategy involves directly fine-tuning \model{Llama2-7B} on parallel documents. The two-stage results are produced by \oursllamasmallfft. In the three-stage strategy, all model parameters of \model{Llama2-7B} are fine-tuned across all three stages.}
\label{tab:training}
\end{table}

\paragraph{Training Strategy}
In this study, we follow the two-stage approach of \citet{DBLP:journals/corr/abs-2309-11674}. Unlike traditional \docnmt methods that typically start with parallel sentence training, we investigate the effectiveness of this strategy on LLM-based \docnmt models. Specifically, we attempt to directly fine-tune the LLMs on parallel documents (one-stage) and add an extra fine-tuning stage using parallel sentences to the training strategy in \autoref{sec:training} (three-stage). The results in \autoref{tab:training} indicate that both the one-stage and three-stage training strategies are sub-optimal for both high-performing languages (Dutch and Romanian) and low-performing languages (Arabic and Chinese) with LLM-based \docnmt models.

% \subsection{Discourse Phenomena}

% \subsection{Scaling Law of Parallel Documents}
% \label{sec:scaling}

\begin{figure}[t]
    \centering
    \begin{tikzpicture}[scale=0.7]
        \begin{axis}[
            xlabel={Percentage (\%) of training data},
            ylabel={\comet},
            ymin=30, ymax=90,
            symbolic x coords = {
            1, 10, 20, 30, 40, 50, 60, 70, 80, 90, 100
            },
            legend pos=south east,
            legend cell align={left},
            legend style={
                legend columns=2,
                font=\small
            },
            width=10cm,
            height=6cm,
            ymajorgrids=true,
            grid style=dashed,
            xtick=data,
        ]
        
        % l-7b-lora
        \addplot[
            color=blue,
            mark=square,
            ]
            coordinates {
            (1,39.82) 
            (10,81.72) 
            (20,82.14) 
            (30,79.02) 
            (40,79.03)
            (50,82.28)
            (60,78.97)
            (70,82.4)
            (80,81.95)
            (90,82.26)
            (100,80.3)
            };
        \addlegendentry{\oursllamasmalllora}

        % l-7b-fft
        \addplot[
            color=cyan,
            mark=square*,
            ]
            coordinates {
            (1,76.88) 
            (10,79.92) 
            (20,71.23) 
            (30,81.15) 
            (40,76.58)
            (50,74.52)
            (60,81.1)
            (70,82.09)
            (80,81.95)
            (90,80.36)
            (100,82.03)
            };
        \addlegendentry{\oursllamasmallfft}

        % b-7b-lora
        \addplot[
            color=green,
            mark=pentagon,
            ]
            coordinates {
            (1,38.21) 
            (10,71.28) 
            (20,72.4) 
            (30,75.32) 
            (40,67.87)
            (50,76.88)
            (60,76.6)
            (70,75.44)
            (80,76.49)
            (90,66.42)
            (100,75.44)
            };
        \addlegendentry{\oursbloomsmalllora}

        % b-7b-fft
        \addplot[
            color=lime,
            mark=pentagon*,
            ]
            coordinates {
            (1,72.44) 
            (10,65.04) 
            (20,72.75) 
            (30,75.88) 
            (40,68.86)
            (50,72.25)
            (60,67.45)
            (70,74.78)
            (80,68.30)
            (90,75.71)
            (100,72.62)
            };
        \addlegendentry{\oursbloomsmallfft}

        % v-7b-lora
        \addplot[
            color=red,
            mark=diamond,
            ]
            coordinates {
            (1,39.87) 
            (10,77.98) 
            (20,70.76) 
            (30,82.11) 
            (40,77.61)
            (50,82.17)
            (60,76.37)
            (70,75.24)
            (80,80.59)
            (90,80.49)
            (100,70.76)
            };
        \addlegendentry{\oursvicunasmalllora}

        % v-7b-fft
        \addplot[
            color=magenta,
            mark=diamond*,
            ]
            coordinates {
            (1,78.08) 
            (10,75.3) 
            (20,77.27) 
            (30,71.95) 
            (40,79.99)
            (50,81.21)
            (60,79.21)
            (70,82.55)
            (80,80.36)
            (90,80.43)
            (100,73.9)
            };
        \addlegendentry{\oursvicunasmallfft}

        \end{axis}
    \end{tikzpicture}
    \caption{\comet-Percentage (\%) of parallel documents for the translations from English to German.}
    \label{fig:scaling_en_de}
\end{figure}
\paragraph{Data Efficiency of Parallel Documents}
We explore the data efficiency of parallel documents. Results for English-German translation are presented in \autoref{fig:scaling_en_de}, and for English-Romanian and English-Chinese in \autoref{sec:scaling_en_ro_zh}. While LLMs typically excel with minimal training data, different fine-tuning strategies show distinct scaling behaviors. Our \model{LoRA} models match full training set performance with just $10\%$ of the data (around $20K$ examples), while fully fine-tuned models achieve near-equivalent performance with only about $1\%$ of the data (approximately $2K$ examples). These are crucial for low-resource languages, as recent LLMs are mainly pre-trained on English text.

% \subsection{Evaluation on Recent Test Sets}
% \label{sec:recent}

\paragraph{Evaluation on Recent Test Sets}
Given their extensive pre-training on large text corpora, LLMs are susceptible to data leakage risks. We evaluate our models using recent test sets from WMT2023 \citep{wmt-2023}. These tests, conducted between English and German, assess the out-of-domain generalization of our models and help mitigate data leakage risks. We use spaCy to segment documents and discard any parallel documents where the source and target sides have a differing number of sentences. Our findings, presented in \autoref{tab:ood_en_de}, reveal that while \mtfive-based models outperform LLM-based models in \autoref{tab:main}, LLM-based models excel in translating out-of-domain text on the WMT2023 test sets. These results highlight the ability of LLM-based \docnmt models to generalize well to out-of-domain translation tasks.

%suggest that the LLM-based \docnmt models are promising for translating out-of-domain text.

\begin{table}[t]
\centering
\small
\setlength{\tabcolsep}{0.5pt}
\begin{tabular}{lcccc}
\toprule
                     & \multicolumn{2}{c}{En-De}     & \multicolumn{2}{c}{De-En}     \\ \cmidrule(rl){2-3} \cmidrule(l){4-5}
                     & \dbleu        & \comet        & \dbleu        & \comet        \\ \midrule
\docnmtmtfivelarge  & 20.2          & 74.4          & 20.0          & 76.5          \\
\model{MR-Doc2Sen-mT5}  & 20.5          & 74.9          & 21.0          & 76.5          \\
\model{MR-Doc2Doc-mT5} & 21.2          & 75.6          & 21.5          & 76.5          \\
\model{DocFlat-mT5}  & 20.9          & 75.1          & 21.8          & 76.5          \\
\model{IADA-mT5}  & 21.2          & 75.4          & 22.0          & 76.5          \\ \midrule
\oursllamasmalllora  & 28.9          & 76.4          & 35.5          & 83.2          \\
\oursllamasmallfft   & \textbf{29.0} & \textbf{77.0} & \textbf{36.1} & \textbf{84.0} \\
\oursbloomsmalllora  & 23.7          & 73.0          & 30.5          & 80.8          \\
\oursbloomsmallfft   & 21.0          & 69.0          & 30.0          & 80.5          \\
\oursvicunasmalllora & 20.5          & 63.8          & 33.9          & 81.8          \\
\oursvicunasmallfft  & 27.8          & 75.0          & 34.7          & 83.1          \\ \bottomrule
\end{tabular}
\caption{
\dbleu and \comet on WMT2023 test sets.
Best results are highlighted in \textbf{bold}.
}
\label{tab:ood_en_de}
\end{table}

% \subsection{Zero-Shot Crosslingual Transfer}
% \label{sec:zero_shot}

% \input{tables/zero_shot_lite}

% \paragraph{Zero-Shot Crosslingual Transfer}
% \label{sec:transfer}

% We also explore the transferability of translation capabilities acquired from one language pair to others. We assess our English-German LLM-based \docnmt models on English-to-other-language test sets, comparing their \comet scores to their base models in \autoref{tab:zero_shot_lite}. Our results indicate that models with fine-tuned instructions (\model{Llama2-7B} and \model{Bloom-7B}) consistently exhibit positive transfer effects across all language pairs, while those with instruction-tuned models (\model{Vicuna-7B}) benefits only few languages. 

% These findings suggest that LLMs are more likely to activate their inherent translation abilities during fine-tuning rather than developing new ones.

% \input{7_discussion}
\section{Conclusion}

% This study investigates the adaptation of LLMs for \docnmt through extensive experimentation. The research involves two fine-tuning methods, three LLM backbones, and 18 translation tasks across nine language pairs. Our findings indicate that while specialized models sometimes surpass GPT-4 in translation accuracy, they continue to face challenges such as off-target translation due to error propagation during decoding. We present a detailed analysis of these LLMs tailored for \docnmt, exploring aspects such as translation errors, discourse phenomena, training strategies, the scaling law of parallel documents, recent test set evaluations, and zero-shot crosslingual transfer. The results highlight both the strengths and limitations of LLM-based \docnmt models, offering a foundation for future research in this field.

In this study, we adapt large language models (LLMs) for document-level machine translation (\docnmt) across specific language pairs and found that moderately-sized, fine-tuned models can sometimes surpass larger models like GPT-4 in translation performance. By investigating prompt strategies and conducting extensive experiments with different fine-tuning methods and LLM backbones, we highlight both the potential and challenges of using LLMs for \docnmt. Despite achieving promising results, issues like off-target translations due to error propagation persist. Our analysis of translation errors, discourse phenomena, strategies for training and inference, and data efficiency underscores the strengths and limitations of LLM-based \docnmt models, providing a foundation for future research to enhance their effectiveness.
\section{Limitations}

\paragraph{Constraints on Model Scale}
Our research is confined to language models of a moderate size, specifically those with $7B$ parameters. This limitation is due to the constraints of our available resources. Consequently, it is crucial to acknowledge that the outcomes of our study might vary if conducted with larger models.

\paragraph{Instability in Training}
The process of supervised fine-tuning for LLMs shows instability in our observations. As detailed in \autoref{fig:scaling_en_de}, there are noticeable inconsistencies in performance. These variations are too significant to attribute solely to the randomness inherent in training. In some cases, the fine-tuning of LLMs fails to reach convergence. Unfortunately, our limited resources restrict us from investigating these failures in depth or devising potential remedies.

\paragraph{Influence of Prompting Techniques}
\autoref{sec:prompting} of our study highlights the significant role of prompting methods in fine-tuning. We experiment with four different prompting techniques. It is important to note that the prompt we recommend may not be the most effective, potentially leading to suboptimal performance of our models.

We acknowledge these limitations and leave them to the future work.

\bibliography{anthology,custom}

\clearpage
\appendix

\begin{table}[t]
\centering
\small
\setlength{\tabcolsep}{0.1pt}
\begin{tabular}{lcccccc}
\toprule
      & \multicolumn{2}{c}{Train} & \multicolumn{2}{c}{Validation} & \multicolumn{2}{c}{Test} \\ \cmidrule(rl){2-3} \cmidrule(rl){4-5} \cmidrule(rl){6-7}
      & \# of sen.  & \# of doc. & \# of sen.    & \# of doc.    & \# of sen. & \# of doc. \\ \midrule
En-Ar & 232K         & 1907       & 2453           & 19            & 1460        & 12         \\
En-De & 206K         & 1705       & 2456           & 19            & 1138        & 10         \\
En-Fr & 233K         & 1914       & 2458           & 19            & 1455        & 12         \\
En-It & 232K         & 1902       & 2495           & 19            & 1147        & 10         \\
En-Ja & 223K         & 1863       & 2420           & 19            & 1452        & 12         \\
En-Ko & 230K         & 1920       & 2437           & 19            & 1429        & 12         \\
En-Nl & 237K         & 1805       & 2780           & 19            & 1181        & 10         \\
En-Ro & 221K         & 1812       & 2592           & 19            & 1129        & 10         \\
En-Zh & 231K         & 1906       & 2436           & 19            & 1459        & 12        \\ \bottomrule
\end{tabular}
\caption{Dataset statistics of parallel documents.}
\label{tab:data_stat}
\end{table}
\section{Statistics of Parallel Documents}
\label{sec:data_stat}

We present the dataset statistics of parallel documents in \autoref{tab:data_stat}.

\section{Optimization and Hyperparameters}
\label{sec:optim}

\paragraph{Fine-tuning on Monolingual Documents}
We fine-tune all the parameters of large language models (LLMs) using a learning rate of $5\times10^{-5}$ and a batch size of $256$. During the training process, we apply the linear learning rate schedule, which includes a warm-up phase comprising $10\%$ of the total training steps.

\paragraph{Fine-tuning on Parallel Documents}
When fine-tuning \oursllamasmalllora and \oursvicunasmalllora on parallel documents, we use a learning rate of $5\times10^{-5}$ and a batch size of $64$. We apply a linear learning rate schedule with a 10\% warm-up phase. The LoRA rank is set to 16, affecting only 0.1\% of the parameters (about 8M parameters). The same hyperparameters are used for fine-tuning \docnmtmtfive models, except for a learning rate of $5\times10^{-4}$. \oursllamasmalllora and \oursvicunasmalllora are fine-tuned for up to $3$ epochs, and \docnmtmtfive models for up to $10$ epochs. Early stopping is based on validation loss.

\begin{figure*}[t]
    \centering
    \begin{subfigure}[t]{0.47\textwidth}
    \centering
    \footnotesize
    \begin{Verbatim}[frame=single, fontsize=\footnotesize, breaklines=true, breakanywhere=true, commandchars=\\\{\}]
[English Context]: And it's truly a great honor to have the opportunity to come to this stage twice; I'm extremely grateful. I have been blown away by this conference, and I want to thank all of you for the many nice comments about what I had to say the other night. And I say that sincerely, partly because  I need that.
[German Context]: Es ist mir wirklich eine Ehre, zweimal auf dieser Bühne stehen zu dürfen. Tausend Dank dafür. Ich bin wirklich begeistert von dieser Konferenz, und ich danke Ihnen allen für die vielen netten Kommentare zu meiner Rede vorgestern Abend. Das meine ich ernst, teilweise deshalb -- weil ich es wirklich brauchen kann!
[English Sentence]: Put yourselves in my position.
[German Sentence]: Versetzen Sie sich mal in meine Lage!

    \end{Verbatim}
    \caption{Prompt 1}
    \label{fig:prompt_example_1}
    \end{subfigure}%
    \hfill
    \begin{subfigure}[t]{0.47\textwidth}
    \centering
    \footnotesize
    \begin{Verbatim}[frame=single, fontsize=\footnotesize, breaklines=true, breakanywhere=true]
[English]: And it's truly a great honor to have the opportunity to come to this stage twice; I'm extremely grateful.
[German]: Es ist mir wirklich eine Ehre, zweimal auf dieser Bühne stehen zu dürfen. Tausend Dank dafür.
[English]: I have been blown away by this conference, and I want to thank all of you for the many nice comments about what I had to say the other night.
[German]: Ich bin wirklich begeistert von dieser Konferenz, und ich danke Ihnen allen für die vielen netten Kommentare zu meiner Rede vorgestern Abend.
[English]: And I say that sincerely, partly because  I need that.
[German]: Das meine ich ernst, teilweise deshalb -- weil ich es wirklich brauchen kann!
[English]: Put yourselves in my position.
[German]: Versetzen Sie sich mal in meine Lage!
    \end{Verbatim}
    \caption{Prompt 2}
    \label{fig:prompt_example_2}
    \end{subfigure}%
    \vfill
    \begin{subfigure}[t]{0.47\textwidth}
    \centering
    \footnotesize
    \begin{Verbatim}[frame=single, fontsize=\footnotesize, breaklines=true, breakanywhere=true, commandchars=\\\{\}]
[English Context]: And it's truly a great honor to have the opportunity to come to this stage twice; I'm extremely grateful. I have been blown away by this conference, and I want to thank all of you for the many nice comments about what I had to say the other night. And I say that sincerely, partly because  I need that.
[German Context]: Es ist mir wirklich eine Ehre, zweimal auf dieser Bühne stehen zu dürfen. Tausend Dank dafür. Ich bin wirklich begeistert von dieser Konferenz, und ich danke Ihnen allen für die vielen netten Kommentare zu meiner Rede vorgestern Abend. Das meine ich ernst, teilweise deshalb -- weil ich es wirklich brauchen kann!
\textcolor{black}{Given the provided parallel context, translate the following English sentence to German:}
[English Sentence]: Put yourselves in my position.
[German Sentence]: Versetzen Sie sich mal in meine Lage!


    \end{Verbatim}
    \caption{Prompt 3}
    \label{fig:prompt_example_3}
    \end{subfigure}%
    \hfill
    \begin{subfigure}[t]{0.47\textwidth}
    \centering
    \footnotesize
    \begin{Verbatim}[frame=single, fontsize=\footnotesize, breaklines=true, breakanywhere=true, commandchars=\\\{\}]
[English]: And it's truly a great honor to have the opportunity to come to this stage twice; I'm extremely grateful.
[German]: Es ist mir wirklich eine Ehre, zweimal auf dieser Bühne stehen zu dürfen. Tausend Dank dafür.
[English]: I have been blown away by this conference, and I want to thank all of you for the many nice comments about what I had to say the other night.
[German]: Ich bin wirklich begeistert von dieser Konferenz, und ich danke Ihnen allen für die vielen netten Kommentare zu meiner Rede vorgestern Abend.
[English]: And I say that sincerely, partly because  I need that.
[German]: Das meine ich ernst, teilweise deshalb -- weil ich es wirklich brauchen kann!
\textcolor{black}{Given the provided parallel sentence pairs, translate the following English sentence to German:}
[English]: Put yourselves in my position.
[German]: Versetzen Sie sich mal in meine Lage!
    \end{Verbatim}
    \caption{Prompt 4}
    \label{fig:prompt_example_4}
    \end{subfigure}%

    \caption{
    Prompt types used in the preliminary study. 
    \texttt{<src\_lang>} and \texttt{<tgt\_lang> indicate the language IDs. \texttt{<src*>} and \texttt{<tgt*>}} indicate the source and target sentences.
    \textbf{Note that the target sentences \texttt{<tgt*>} are only used during training and are replaced with the hypotheses \texttt{<hyp*>} generated by the model during inference.}
    }
    \label{fig:prompt_example}
\end{figure*}
\section{Prompt Types}
\label{sec:prompt_types}
We present concrete examples of prompt variations in \autoref{fig:prompt_example}.

\begin{figure*}[t]
    \centering
    \footnotesize
    \begin{Verbatim}[frame=single, fontsize=\footnotesize, breaklines=true, breakanywhere=true, commandchars=\\\{\}]
[Context]:

[Source]: <src1>
[Reference]: <tgt1>
[Hypothesis]: <hyp1>
[Source]: <src2>
[Reference]: <tgt2>
[Hypothesis]: <hyp2>
[Source]: <src3>
[Reference]: <tgt3>
[Hypothesis]: <hyp3>

[Current Sentence]:

[Source]: <src4>
[Reference]: <tgt4>
[Hypothesis]: <hyp4>

[Error Types]:

- Mistranslation: Error occuring when the target content does not accurately represent the source content.
- Overtranslation: Error occuring in the target content that is inappropriately more specific than the source content.
- Undertranslation: Error occuring in the target content that is inappropriately less specific than the source content.
- Addition: Error occuring in the target content that includes content not present in the source.
- Omission: Error where content present in the source is missing in the target.
- Unjustfied euphemism: Target content that is potentially offensive in some way in the source language, but that has been inappropriately "watered down" in the translation.
- Do not translate: Error occuring when a text segment marked "Do not translate!" is translated in the target text.
- Untranslated: Error occuring when a text segment that was intended for translation is omitted in the target content.
- Retained factual error: Untrue statement or an incorrect data value present in the source content and retained in the target content.
- Completeness: Source text incomplete, resulting in instances where needed content is missing in the source language.
- Grammar: Error that occurs when a text string (sentence, phrase, other) in the translation violates the grammatical rules of the target language.
- Punctuation: Punctuation incorrect according to target language conventions.
- Spelling: Error occurring when a word is misspelled.
- Duplication: Content (e.g., a word or longer portion of text) repeated unintentionally.
- Unclear reference: Relative pronouns or other referential mechanisms unclear in their reference.
- Cohesion: Portions of the text needed to connect it into an understandable whole (e.g., reference, substitution, ellipsis, conjunction, and lexical cohesion) missing or incorrect.
- Coherence: Text lacking a clear semantic relationship between its parts, i.e., the different parts don't hang together, don't follow the discourse conventions of the target language, or don't "make sense."
- Inconsistent style: Style that varies inconsistently throughout the text, e.g., One part of a text is written in a clear, "terse" style, while other sections are written in a more wordy style.
- Multiple terms in translation: Error where source content terminology is correct, but target content terms are not used consistently.

Considering the provided context, please identify the errors of the translation from the source to the target in the current sentence based on a subset of Multidimensional Quality Metrics (MQM) error typology.
You should pay extra attention to the error types related to the relationship between the current sentence and its context, such as "Unclear reference", "Cohesion", "Coherence", "Inconsistent style", and "Multiple terms in translation".
You should list all the errors you find in the sentence, and provide a justification for each error.
Your output should always be in JSON format, formatted as follows: \{'justification': '...', 'error_types': [...]\}.
    \end{Verbatim}
    \caption{Prompt used for analyzing translation error types.}
    \label{fig:error_type_prompt}
\end{figure*}
\section{GPT-4 Prompts}
\label{sec:gpt_prompts}
We present the prompts used for error type analysis in \autoref{fig:error_type_prompt}.

\begin{table*}[t]
\centering
\small
\begin{tabular}{lcccccccccc}
\toprule
                     & \avgsbleu      & Ar             & De             & Fr             & It             & Ja             & Ko             & Nl             & Ro             & Zh             \\ \midrule
\multicolumn{11}{l}{\cellcolor{gray!30}\textit{State-of-the-art \sennmt baselines}}                                                                                                                               \\
\nllbsmall           & 23.6           & 14.2           & 22.2           & 38.5           & 36.0           & 11.4           & 16.2           & 30.2           & 25.6           & 17.6           \\
\nllbmedium          & 25.7           & 16.2           & 27.6           & 40.6           & 37.7           & 12.6           & 18.6           & 32.2           & 27.3           & 18.3           \\
\nllbbig             & 26.8           & 17.4           & 28.8           & 41.3           & 39.2           & 14.1           & 19.5           & 33.7           & 28.1           & 18.7           \\
\googletrans         & 24.5           & 14.2           & 25.3           & 38.0           & 35.0           & 11.6           & 16.5           & 29.6           & 24.0           & 26.4           \\ \midrule
\multicolumn{11}{l}{\cellcolor{gray!30}\textit{State-of-the-art LLMs}}                                                                                                                                            \\
\chatgpt             & 26.3           & 14.9           & 27.2           & 40.5           & 36.6           & 13.2           & 15.9           & 31.5           & 26.6           & 30.4           \\
\gptfour             & 27.0           & 16.1           & 27.4           & 40.0           & 35.8           & 14.1           & 18.3           & 32.2           & 27.3           & 31.6           \\ \midrule
\multicolumn{11}{l}{\cellcolor{gray!30}\textit{LLM backbones}}                                                                                                                                                    \\
\model{Llama2}       & \phantom{0}2.8 & \phantom{0}0.4 & \phantom{0}6.6 & \phantom{0}4.5 & \phantom{0}1.0 & \phantom{0}1.0 & \phantom{0}1.9 & \phantom{0}0.2 & \phantom{0}1.9 & \phantom{0}7.4 \\
\model{Bloom}        & \phantom{0}2.5 & \phantom{0}1.0 & \phantom{0}1.0 & 12.1           & \phantom{0}1.4 & \phantom{0}0.1 & \phantom{0}3.1 & \phantom{0}0.7 & \phantom{0}0.1 & \phantom{0}3.4 \\
\model{Vicuna}       & 10.2           & \phantom{0}4.5 & \phantom{0}6.4 & \phantom{0}6.4 & \phantom{0}8.6 & 10.2           & \phantom{0}9.8 & 13.9           & \phantom{0}6.8 & 25.4           \\ \midrule
\multicolumn{11}{l}{\cellcolor{gray!30}\textit{Re-implemented \docnmt baselines}}                                                                                                                                 \\
\docnmtmtfivesmall   & 17.2           & \phantom{0}9.4 & 16.8           & 24.0           & 21.0           & 11.0           & 13.7           & 20.5           & 17.1           & 21.6           \\
\docnmtmtfivebase    & 18.6           & 10.8           & 18.2           & 24.9           & 23.0           & 12.9           & 15.2           & 21.7           & 17.8           & 22.9           \\
\docnmtmtfivelarge   & 18.4           & 10.3           & 18.1           & 24.9           & 22.4           & 13.9           & 15.4           & 19.6           & 18.8           & 22.6           \\
\mrdoctosent         & 18.8           & 10.2           & 18.8           & 25.6           & 22.3           & 14.5           & 16.2           & 19.6           & 19.3           & 22.8           \\
\mrdoctodoc          & ---            & ---            & ---            & ---            & ---            & ---            & ---            & ---            & ---            & ---            \\
\docflat             & 19.2           & 11.0           & 19.2           & 25.7           & 22.6           & 14.7           & 16.5           & 20.3           & 19.2           & 23.8           \\
\iada                & 19.7           & 11.7           & 19.4           & 26.3           & 23.9           & 15.2           & 16.9           & 20.9           & 19.6           & 23.4           \\ \midrule
\multicolumn{11}{l}{\cellcolor{gray!30}\textit{Bilingual English-from/to-Many LLM-based \docnmt models (Ours)}}                                                                                                                        \\
\oursllamasmalllora  & 17.2           & 13.0           & 25.1           & 34.9           & \phantom{0}6.8 & \phantom{0}8.7 & 13.0           & \phantom{0}3.7 & 22.7           & 27.3           \\
\oursllamasmallfft   & 13.7           & 13.1           & 25.3           & 19.5           & \phantom{0}2.6 & \phantom{0}7.9 & \phantom{0}7.2 & \phantom{0}4.1 & 21.1           & 22.8           \\
\oursbloomsmalllora  & 17.7           & 12.1           & 20.6           & 32.6           & 32.9           & \phantom{0}3.6 & \phantom{0}1.4 & 28.1           & 12.2           & 15.7           \\
\oursbloomsmallfft   & 12.0           & 10.1           & 19.6           & 38.5           & \phantom{0}0.1 & \phantom{0}1.9 & \phantom{0}2.4 & \phantom{0}1.5 & 19.9           & 14.5           \\
\oursvicunasmalllora & 16.4           & 13.3           & 20.1           & 20.7           & 13.6           & \phantom{0}9.1 & 14.3           & \phantom{0}5.5 & 23.0           & 28.1           \\
\oursvicunasmallfft  & 14.3           & 13.5           & 23.3           & 21.1           & \phantom{0}4.8 & \phantom{0}3.8 & 15.9           & \phantom{0}3.3 & 17.4           & 25.8           \\ \midrule
\multicolumn{11}{l}{\cellcolor{gray!30}\textit{Multilingual English-from/to-Many LLM-based \docnmt models (Ours)}}                                                                                                             \\
\oursllamasmalllora  & 13.9           & \phantom{0}8.8 & 22.2           & 27.9           & \phantom{0}5.0 & \phantom{0}7.7 & \phantom{0}9.4 & \phantom{0}3.4 & 19.4           & 21.6           \\
\oursllamasmallfft   & 17.1           & \phantom{0}7.8 & 22.0           & 32.5           & 17.9           & 12.5           & 11.5           & \phantom{0}4.2 & 20.3           & 25.1           \\
\oursbloomsmalllora  & 10.3           & 12.5           & \phantom{0}3.8 & 36.0           & 12.8           & \phantom{0}5.8 & \phantom{0}1.6 & \phantom{0}3.7 & \phantom{0}3.4 & 12.9           \\
\oursbloomsmallfft   & 16.0           & 12.6           & 15.8           & 37.4           & 27.7           & \phantom{0}2.0 & \phantom{0}1.3 & 18.4           & \phantom{0}7.8 & 21.3           \\
\oursvicunasmalllora & 12.9           & \phantom{0}8.0 & 15.4           & 28.2           & \phantom{0}6.2 & 10.1           & \phantom{0}7.1 & \phantom{0}3.3 & 15.1           & 22.4           \\
\oursvicunasmallfft  & 13.8           & 10.4           & 15.2           & 11.3           & \phantom{0}8.4 & 11.1           & 12.2           & 12.4           & 17.0           & 26.3           \\ \bottomrule
\end{tabular}
\caption{Breakdown \sbleu results for the translation tasks from English to other languages.}
\label{tab:sbleu_en_x}
\end{table*}
\begin{table*}[t]
\centering
\small
\begin{tabular}{lcccccccccc}
\toprule
                     & \avgdbleu      & Ar             & De             & Fr             & It             & Ja             & Ko             & Nl             & Ro             & Zh             \\ \midrule
\multicolumn{11}{l}{\cellcolor{gray!30}\textit{State-of-the-art \sennmt baselines}}                                                                                                                               \\
\nllbsmall           & 27.3           & 15.4           & 26.0           & 42.1           & 39.0           & 17.1           & 23.8           & 33.8           & 28.5           & 20.2           \\
\nllbmedium          & 29.5           & 17.4           & 31.8           & 44.0           & 40.8           & 18.1           & 26.6           & 35.8           & 30.2           & 21.0           \\
\nllbbig             & 30.5           & 18.6           & 32.9           & 44.6           & 42.2           & 19.8           & 27.4           & 37.0           & 30.9           & 21.2           \\
\googletrans         & 28.4           & 16.0           & 29.3           & 41.3           & 38.5           & 15.7           & 23.4           & 32.8           & 26.7           & 32.1           \\
\multicolumn{11}{l}{\cellcolor{gray!30}\textit{State-of-the-art LLMs}}                                                                                                                                            \\
\chatgpt             & 30.1           & 16.4           & 30.9           & 43.7           & 39.7           & 17.5           & 22.7           & 34.5           & 29.0           & 36.3           \\
\gptfour             & 30.7           & 17.4           & 31.1           & 43.2           & 39.0           & 18.4           & 25.3           & 35.3           & 29.8           & 37.2           \\ \midrule
\multicolumn{11}{l}{\cellcolor{gray!30}\textit{LLM backbones}}                                                                                                                                                    \\
\model{Llama2}       & \phantom{0}3.5 & \phantom{0}0.5 & \phantom{0}7.4 & \phantom{0}4.9 & \phantom{0}1.1 & \phantom{0}1.8 & \phantom{0}3.7 & \phantom{0}0.2 & \phantom{0}2.2 & \phantom{0}9.6 \\
\model{Bloom}        & \phantom{0}2.8 & \phantom{0}1.0 & \phantom{0}1.3 & 12.9           & \phantom{0}1.7 & \phantom{0}0.3 & \phantom{0}2.7 & \phantom{0}1.0 & \phantom{0}0.1 & \phantom{0}4.4 \\
\model{Vicuna}       & 12.4           & \phantom{0}5.7 & \phantom{0}6.4 & \phantom{0}7.1 & \phantom{0}8.5 & 15.0           & 16.1           & 14.4           & \phantom{0}7.4 & 31.2           \\ \midrule
\multicolumn{11}{l}{\cellcolor{gray!30}\textit{Re-implemented \docnmt baselines}}                                                                                                                                 \\
\docnmtmtfivesmall   & 20.2           & 10.3           & 18.8           & 26.1           & 21.9           & 16.8           & 21.5           & 21.5           & 18.4           & 26.6           \\
\docnmtmtfivebase    & 21.5           & 11.7           & 20.0           & 27.0           & 23.9           & 18.4           & 23.2           & 22.6           & 18.8           & 28.0           \\
\docnmtmtfivelarge   & 21.4           & 11.2           & 20.1           & 27.1           & 23.4           & 19.7           & 23.0           & 20.7           & 20.2           & 27.1           \\
\mrdoctosent         & ---            & ---            & ---            & ---            & ---            & ---            & ---            & ---            & ---            & ---            \\
\mrdoctodoc          & 21.9           & 11.9           & 20.7           & 27.9           & 23.8           & 19.8           & 23.3           & 21.5           & 20.7           & 27.9           \\
\docflat             & 22.5           & 12.1           & 20.8           & 28.0           & 24.7           & 20.9           & 24.3           & 21.9           & 21.7           & 27.9           \\
\iada                & 22.4           & 12.2           & 21.3           & 28.4           & 24.0           & 20.8           & 24.1           & 21.1           & 21.4           & 28.1           \\ \midrule
\multicolumn{11}{l}{\cellcolor{gray!30}\textit{Bilingual English-from/to-Many LLM-based \docnmt models (Ours)}}                                                                                                                        \\
\oursllamasmalllora  & 20.2           & 14.7           & 29.1           & 37.5           & \phantom{0}7.3 & 13.9           & 19.5           & \phantom{0}4.2 & 22.9           & 33.1           \\
\oursllamasmallfft   & 16.2           & 14.7           & 29.4           & 20.6           & \phantom{0}2.7 & 12.5           & 12.3           & \phantom{0}4.5 & 21.6           & 27.6           \\
\oursbloomsmalllora  & 20.5           & 13.7           & 24.8           & 36.1           & 36.3           & \phantom{0}6.9 & \phantom{0}2.6 & 32.2           & 12.2           & 19.7           \\
\oursbloomsmallfft   & 13.8           & 11.2           & 23.6           & 41.7           & \phantom{0}0.1 & \phantom{0}3.7 & \phantom{0}3.9 & \phantom{0}1.7 & 20.1           & 18.1           \\
\oursvicunasmalllora & 19.3           & 14.9           & 23.1           & 21.8           & 14.7           & 14.3           & 21.6           & \phantom{0}5.9 & 23.3           & 34.2           \\
\oursvicunasmallfft  & 16.8           & 15.2           & 26.9           & 22.3           & \phantom{0}4.9 & \phantom{0}6.2 & 23.6           & \phantom{0}3.7 & 17.5           & 31.1           \\ \midrule
\multicolumn{11}{l}{\cellcolor{gray!30}\textit{Multilingual English-from/to-Many LLM-based \docnmt models (Ours)}}                                                                                                             \\
\oursllamasmalllora  & 16.9           & 10.2           & 26.0           & 30.4           & \phantom{0}5.6 & 12.8           & 15.7           & \phantom{0}3.8 & 20.2           & 27.6           \\
\oursllamasmallfft   & 20.4           & \phantom{0}9.2 & 25.5           & 35.5           & 18.5           & 19.2           & 18.3           & \phantom{0}4.9 & 20.9           & 31.6           \\
\oursbloomsmalllora  & 12.3           & 14.3           & \phantom{0}4.4 & 39.5           & 14.4           & 10.8           & \phantom{0}3.3 & \phantom{0}3.9 & \phantom{0}3.5 & 16.8           \\
\oursbloomsmallfft   & 18.6           & 14.2           & 19.5           & 40.9           & 30.9           & \phantom{0}3.8 & \phantom{0}2.6 & 20.9           & \phantom{0}8.5 & 26.5           \\
\oursvicunasmalllora & 15.7           & \phantom{0}9.5 & 18.1           & 31.0           & \phantom{0}6.5 & 16.0           & 11.8           & \phantom{0}3.9 & 16.1           & 28.1           \\
\oursvicunasmallfft  & 16.8           & 12.0           & 17.5           & 12.9           & \phantom{0}8.8 & 17.4           & 19.5           & 12.4           & 17.8           & 32.7           \\ \bottomrule
\end{tabular}
\caption{Breakdown \dbleu results for the translation tasks from English to other languages.}
\label{tab:dbleu_en_x}
\end{table*}
\begin{table*}[t]
\centering
\small
\begin{tabular}{lcccccccccc}
\toprule
                     & \avgcomet & Ar   & De   & Fr   & It   & Ja   & Ko   & Nl   & Ro   & Zh   \\ \midrule
\multicolumn{11}{l}{\cellcolor{gray!30}\textit{State-of-the-art \sennmt baselines}}                                \\
\nllbsmall           & 82.3      & 82.6 & 81.3 & 83.7 & 86.3 & 79.5 & 82.5 & 84.0 & 85.0 & 75.4 \\
\nllbmedium          & 83.5      & 84.3 & 83.0 & 84.8 & 87.5 & 80.1 & 84.2 & 85.2 & 86.3 & 76.4 \\
\nllbbig             & 84.3      & 84.8 & 84.1 & 85.3 & 88.0 & 82.1 & 85.2 & 86.0 & 86.4 & 76.7 \\
\googletrans         & 81.6      & 81.5 & 80.2 & 82.3 & 85.1 & 80.7 & 79.9 & 83.7 & 82.9 & 78.5 \\ \midrule
\multicolumn{11}{l}{\cellcolor{gray!30}\textit{State-of-the-art LLMs}}                                             \\
\chatgpt             & 85.3      & 83.8 & 84.6 & 85.9 & 87.7 & 84.7 & 84.2 & 86.3 & 86.5 & 83.9 \\
\gptfour             & 86.3      & 85.4 & 85.4 & 86.2 & 87.8 & 86.0 & 86.4 & 87.0 & 87.4 & 84.8 \\ \midrule
\multicolumn{11}{l}{\cellcolor{gray!30}\textit{LLM backbones}}                                                     \\
\model{Llama2}       & 40.1      & 37.4 & 41.5 & 41.2 & 39.3 & 39.7 & 42.8 & 35.5 & 41.6 & 42.1 \\
\model{Bloom}        & 35.5      & 34.9 & 34.8 & 45.3 & 33.0 & 33.9 & 35.5 & 34.2 & 28.6 & 39.0 \\
\model{Vicuna}       & 64.7      & 68.3 & 48.7 & 49.0 & 62.5 & 81.3 & 72.4 & 64.1 & 56.1 & 80.2 \\ \midrule
\multicolumn{11}{l}{\cellcolor{gray!30}\textit{Re-implemented \docnmt baselines}}                                  \\
\docnmtmtfivesmall   & 75.1      & 77.2 & 71.0 & 72.4 & 74.1 & 78.5 & 77.8 & 74.3 & 74.0 & 77.0 \\
\docnmtmtfivebase    & 78.3      & 80.8 & 74.4 & 74.8 & 77.9 & 81.5 & 82.0 & 76.8 & 76.6 & 79.6 \\
\docnmtmtfivelarge   & 79.2      & 81.1 & 75.9 & 75.9 & 78.9 & 82.9 & 82.4 & 76.5 & 79.3 & 80.3 \\
\mrdoctosent         & 79.9      & 82.2 & 76.3 & 76.5 & 80.0 & 83.6 & 83.1 & 76.7 & 79.7 & 80.9 \\
\mrdoctodoc          & ---       & ---  & ---  & ---  & ---  & ---  & ---  & ---  & ---  & ---  \\
\docflat             & 80.4      & 81.8 & 77.3 & 76.9 & 80.3 & 83.6 & 83.4 & 78.0 & 80.7 & 81.8 \\
\iada                & 80.7      & 82.4 & 77.0 & 77.3 & 80.9 & 84.1 & 83.7 & 77.8 & 80.9 & 81.8 \\ \midrule
\multicolumn{11}{l}{\cellcolor{gray!30}\textit{Bilingual English-from/to-Many LLM-based \docnmt models (Ours)}}                         \\
\oursllamasmalllora  & 70.8      & 82.5 & 80.3 & 79.1 & 42.9 & 70.4 & 75.4 & 42.5 & 83.6 & 80.6 \\
\oursllamasmallfft   & 67.4      & 83.1 & 82.0 & 59.4 & 38.8 & 65.8 & 69.7 & 49.2 & 82.7 & 75.5 \\
\oursbloomsmalllora  & 68.5      & 77.0 & 75.4 & 76.8 & 85.1 & 51.4 & 40.4 & 82.9 & 61.7 & 65.5 \\
\oursbloomsmallfft   & 59.6      & 68.4 & 72.6 & 83.8 & 45.3 & 40.3 & 45.2 & 46.8 & 71.6 & 62.2 \\
\oursvicunasmalllora & 69.7      & 82.7 & 70.8 & 60.1 & 56.2 & 70.2 & 76.9 & 44.1 & 85.0 & 81.3 \\
\oursvicunasmallfft  & 65.0      & 83.1 & 73.9 & 58.8 & 41.5 & 54.8 & 81.1 & 42.4 & 69.6 & 79.4 \\ \midrule
\multicolumn{11}{l}{\cellcolor{gray!30}\textit{Multilingual English-from/to-Many LLM-based \docnmt models (Ours)}}              \\
\oursllamasmalllora  & 67.2      & 78.8 & 74.1 & 69.3 & 40.9 & 65.9 & 74.7 & 44.9 & 79.3 & 77.1 \\
\oursllamasmallfft   & 73.8      & 78.8 & 76.3 & 77.0 & 65.2 & 80.6 & 79.0 & 45.3 & 81.9 & 79.8 \\
\oursbloomsmalllora  & 58.1      & 80.9 & 43.8 & 79.9 & 54.9 & 66.7 & 46.6 & 38.1 & 46.9 & 65.3 \\
\oursbloomsmallfft   & 67.8      & 82.2 & 69.6 & 83.8 & 81.7 & 46.3 & 45.7 & 70.5 & 54.5 & 76.0 \\
\oursvicunasmalllora & 66.4      & 78.2 & 65.5 & 70.8 & 46.9 & 74.5 & 66.4 & 41.9 & 76.0 & 77.6 \\
\oursvicunasmallfft  & 69.3      & 80.3 & 64.7 & 54.2 & 52.0 & 78.1 & 77.5 & 57.6 & 78.1 & 81.1 \\ \bottomrule
\end{tabular}
\caption{Breakdown \comet results for the translation tasks from English to other languages.}
\label{tab:comet_en_x}
\end{table*}

\begin{table*}[t]
\centering
\small
\begin{tabular}{lcccccccccc}
\toprule
                     & \avgsbleu      & Ar             & De             & Fr             & It             & Ja             & Ko             & Nl             & Ro             & Zh             \\ \midrule
\multicolumn{11}{l}{\cellcolor{gray!30}\textit{State-of-the-art \sennmt baselines}}                                                                                                                               \\
\nllbsmall           & 18.2           & 26.8           & 11.0           & 31.0           & 22.7           & 10.9           & 13.6           & 17.1           & 13.2           & 17.9           \\
\nllbmedium          & 25.0           & 35.9           & 18.8           & 37.4           & 35.3           & 13.2           & 15.7           & 26.0           & 23.1           & 19.9           \\
\nllbbig             & 25.8           & 36.5           & 22.3           & 36.8           & 33.5           & 12.4           & 18.5           & 28.3           & 25.2           & 19.1           \\
\googletrans         & 25.0           & 28.7           & 26.1           & 34.7           & 35.1           & 10.2           & 13.3           & 30.8           & 29.6           & 16.6           \\ \midrule
\multicolumn{11}{l}{\cellcolor{gray!30}\textit{State-of-the-art LLMs}}                                                                                                                                            \\
\chatgpt             & 30.7           & 35.8           & 30.8           & 40.7           & 41.8           & 15.5           & 17.3           & 36.1           & 35.4           & 22.9           \\
\gptfour             & 31.7           & 37.2           & 31.2           & 41.4           & 42.3           & 15.9           & 19.8           & 36.6           & 36.5           & 24.4           \\ \midrule
\multicolumn{11}{l}{\cellcolor{gray!30}\textit{LLM backbones}}                                                                                                                                                    \\
\model{Llama2}       & \phantom{0}4.2 & \phantom{0}0.1 & \phantom{0}6.2 & \phantom{0}1.3 & \phantom{0}4.4 & \phantom{0}0.0 & \phantom{0}0.1 & 15.1           & 10.3           & \phantom{0}0.1 \\
\model{Bloom}        & \phantom{0}6.7 & \phantom{0}4.7 & \phantom{0}8.7 & 14.3           & 16.9           & \phantom{0}0.1 & \phantom{0}0.3 & \phantom{0}9.4 & \phantom{0}5.5 & \phantom{0}0.2 \\
\model{Vicuna}       & \phantom{0}9.5 & \phantom{0}1.1 & 14.4           & 24.9           & 14.1           & \phantom{0}4.7 & \phantom{0}0.4 & 17.2           & \phantom{0}8.5 & \phantom{0}0.0 \\ \midrule
\multicolumn{11}{l}{\cellcolor{gray!30}\textit{Re-implemented \docnmt baselines}}                                                                                                                                 \\
\docnmtmtfivesmall   & 19.4           & 23.0           & 19.5           & 26.6           & 25.4           & \phantom{0}9.5 & 11.4           & 22.0           & 22.6           & 14.5           \\
\docnmtmtfivebase    & 20.7           & 24.2           & 20.3           & 28.0           & 26.6           & 11.0           & 11.6           & 24.4           & 23.8           & 16.1           \\
\docnmtmtfivelarge   & 21.5           & 25.7           & 21.0           & 28.7           & 27.3           & 11.0           & 12.8           & 25.4           & 25.0           & 16.8           \\
\mrdoctosent         & 22.5           & 26.9           & 22.0           & 29.9           & 27.7           & 11.8           & 13.9           & 26.5           & 26.1           & 18.0           \\
\mrdoctodoc          & ---            & ---            & ---            & ---            & ---            & ---            & ---            & ---            & ---            & ---            \\
\docflat             & 22.5           & 26.6           & 22.3           & 29.7           & 28.4           & 11.7           & 13.9           & 26.4           & 25.9           & 17.6           \\
\iada                & 23.1           & 26.9           & 22.7           & 30.5           & 28.5           & 12.8           & 14.5           & 27.1           & 26.1           & 18.7           \\ \midrule
\multicolumn{11}{l}{\cellcolor{gray!30}\textit{Multilingual English-from/to-Many LLM-based \docnmt models (Ours)}}                                                                                                                        \\
\oursllamasmalllora  & 23.8           & \phantom{0}3.9 & 33.1           & 40.3           & 45.2           & \phantom{0}8.3 & \phantom{0}5.0 & 39.2           & 39.0           & \phantom{0}0.1 \\
\oursllamasmallfft   & 22.4           & \phantom{0}2.5 & 32.2           & 42.6           & 44.8           & \phantom{0}1.0 & \phantom{0}1.0 & 38.9           & 38.2           & \phantom{0}0.1 \\
\oursbloomsmalllora  & 29.9           & 30.9           & 30.5           & 41.1           & 41.2           & 13.9           & 15.5           & 35.0           & 35.2           & 25.6           \\
\oursbloomsmallfft   & 22.3           & 17.0           & 29.8           & 41.3           & 40.7           & \phantom{0}0.4 & \phantom{0}1.1 & 35.6           & 34.3           & \phantom{0}1.0 \\
\oursvicunasmalllora & 21.6           & \phantom{0}3.8 & 30.8           & 43.1           & 29.4           & \phantom{0}5.5 & \phantom{0}4.0 & 39.1           & 38.7           & \phantom{0}0.3 \\
\oursvicunasmallfft  & 24.3           & 24.2           & 31.0           & 43.4           & 45.0           & \phantom{0}0.0 & \phantom{0}0.7 & 36.2           & 38.0           & \phantom{0}0.1 \\ \midrule
\multicolumn{11}{l}{\cellcolor{gray!30}\textit{Multilingual English-from/to-Many LLM-based \docnmt models (Ours)}}                                                                                                             \\
\oursllamasmalllora  & 17.1           & \phantom{0}2.3 & 15.7           & 18.6           & 41.6           & \phantom{0}0.4 & \phantom{0}0.8 & 37.6           & 36.7           & \phantom{0}0.3 \\
\oursllamasmallfft   & 18.3           & 15.8           & 10.3           & 22.0           & 40.7           & \phantom{0}2.1 & \phantom{0}6.0 & 29.7           & 30.8           & \phantom{0}7.8 \\
\oursbloomsmalllora  & 23.5           & 34.4           & 28.5           & 39.5           & 40.0           & \phantom{0}0.1 & \phantom{0}0.6 & 33.7           & 34.1           & \phantom{0}0.6 \\
\oursbloomsmallfft   & 27.2           & 35.9           & 28.3           & 39.7           & 38.8           & \phantom{0}7.1 & 11.2           & 32.4           & 33.5           & 17.9           \\
\oursvicunasmalllora & 17.2           & \phantom{0}6.1 & \phantom{0}5.5 & 12.2           & 43.3           & \phantom{0}2.9 & \phantom{0}1.8 & 35.9           & 36.4           & 10.2           \\
\oursvicunasmallfft  & 18.4           & 13.0           & 17.7           & 16.7           & 37.6           & \phantom{0}2.5 & \phantom{0}3.7 & 36.2           & 35.2           & \phantom{0}3.0 \\ \bottomrule
\end{tabular}
\caption{Breakdown \sbleu results for the translation tasks from other languages to English.}
\label{tab:sbleu_x_en}
\end{table*}
\begin{table*}[t]
\centering
\small
\begin{tabular}{lcccccccccc}
\toprule
                     & \avgdbleu      & Ar             & De             & Fr             & It             & Ja             & Ko             & Nl   & Ro             & Zh             \\ \midrule
\multicolumn{11}{l}{\cellcolor{gray!30}\textit{State-of-the-art \sennmt baselines}}                                                                                                                     \\
\nllbsmall           & 22.0           & 30.5           & 14.7           & 34.1           & 26.3           & 14.8           & 18.1           & 21.0 & 16.8           & 22.2           \\
\nllbmedium          & 28.6           & 39.2           & 22.6           & 40.1           & 38.7           & 17.1           & 20.3           & 29.4 & 26.7           & 23.7           \\
\nllbbig             & 29.4           & 39.7           & 26.1           & 39.6           & 37.0           & 16.5           & 23.2           & 31.3 & 28.8           & 22.8           \\
\googletrans         & 28.5           & 32.0           & 29.8           & 37.8           & 38.9           & 13.3           & 17.7           & 33.7 & 33.1           & 20.4           \\
\multicolumn{11}{l}{\cellcolor{gray!30}\textit{State-of-the-art LLMs}}                                                                                                                                  \\
\chatgpt             & 34.0           & 38.8           & 34.0           & 43.4           & 44.8           & 19.5           & 22.0           & 38.8 & 38.4           & 26.9           \\
\gptfour             & 35.1           & 40.2           & 34.5           & 44.2           & 46.3           & 19.7           & 24.4           & 39.3 & 39.6           & 28.1           \\ \midrule
\multicolumn{11}{l}{\cellcolor{gray!30}\textit{LLM backbones}}                                                                                                                                          \\
\model{Llama2}       & \phantom{0}4.4 & \phantom{0}0.1 & \phantom{0}6.9 & \phantom{0}1.5 & \phantom{0}4.7 & \phantom{0}0.0 & \phantom{0}0.1 & 15.7 & 10.9           & \phantom{0}0.1 \\
\model{Bloom}        & \phantom{0}7.3 & \phantom{0}5.3 & \phantom{0}9.8 & 15.2           & 17.6           & \phantom{0}0.1 & \phantom{0}0.5 & 10.6 & \phantom{0}6.6 & \phantom{0}0.2 \\
\model{Vicuna}       & \phantom{0}9.8 & \phantom{0}1.1 & 14.5           & 24.6           & 14.4           & \phantom{0}5.9 & \phantom{0}0.5 & 18.5 & \phantom{0}8.9 & \phantom{0}0.0 \\ \midrule
\multicolumn{11}{l}{\cellcolor{gray!30}\textit{Re-implemented \docnmt baselines}}                                                                                                                       \\
\docnmtmtfivesmall   & 21.2           & 24.5           & 21.1           & 27.5           & 26.5           & 12.6           & 14.1           & 23.5 & 23.9           & 17.0           \\
\docnmtmtfivebase    & 22.5           & 25.5           & 22.1           & 28.9           & 27.8           & 14.0           & 14.6           & 25.8 & 25.2           & 18.5           \\
\docnmtmtfivelarge   & 23.4           & 26.9           & 23.0           & 29.7           & 28.4           & 14.2           & 15.7           & 26.8 & 26.3           & 19.4           \\
\mrdoctosent         & 24.0           & 27.4           & 24.2           & 30.3           & 29.4           & 14.9           & 16.1           & 27.5 & 26.8           & 19.8           \\
\mrdoctodoc          & 24.6           & 28.3           & 24.3           & 30.5           & 29.8           & 15.7           & 16.8           & 27.8 & 27.8           & 20.8           \\
\docflat             & 24.6           & 27.6           & 24.5           & 31.1           & 29.7           & 15.1           & 17.0           & 28.1 & 27.8           & 20.3           \\
\iada                & 24.5           & 28.2           & 24.6           & 30.9           & 29.6           & 15.0           & 17.1           & 27.8 & 27.1           & 20.5           \\ \midrule
\multicolumn{11}{l}{\cellcolor{gray!30}\textit{Bilingual English-from/to-Many LLM-based \docnmt models (Ours)}}                                                                                                              \\
\oursllamasmalllora  & 25.7           & \phantom{0}4.1 & 36.2           & 42.2           & 48.5           & 10.0           & \phantom{0}5.9 & 42.2 & 42.1           & \phantom{0}0.1 \\
\oursllamasmallfft   & 24.1           & \phantom{0}2.6 & 35.3           & 45.1           & 48.1           & \phantom{0}1.0 & \phantom{0}1.1 & 41.9 & 41.4           & \phantom{0}0.1 \\
\oursbloomsmalllora  & 33.6           & 33.0           & 34.2           & 44.0           & 44.9           & 18.8           & 20.4           & 38.1 & 38.6           & 30.4           \\
\oursbloomsmallfft   & 24.5           & 18.6           & 33.4           & 44.2           & 44.4           & \phantom{0}0.6 & \phantom{0}1.4 & 38.7 & 37.9           & \phantom{0}1.1 \\
\oursvicunasmalllora & 23.3           & \phantom{0}3.8 & 33.9           & 45.5           & 30.6           & \phantom{0}6.7 & \phantom{0}4.8 & 42.1 & 42.0           & \phantom{0}0.3 \\
\oursvicunasmallfft  & 23.5           & \phantom{0}2.3 & 34.1           & 46.0           & 48.3           & \phantom{0}0.0 & \phantom{0}0.7 & 39.2 & 41.0           & \phantom{0}0.0 \\ \midrule
\multicolumn{11}{l}{\cellcolor{gray!30}\textit{Multilingual English-from/to-Many LLM-based \docnmt models (Ours)}}                                                                                                   \\
\oursllamasmalllora  & 18.4           & \phantom{0}2.2 & 17.6           & 19.5           & 44.6           & \phantom{0}0.4 & \phantom{0}0.8 & 40.4 & 39.7           & \phantom{0}0.2 \\
\oursllamasmallfft   & 19.5           & 15.4           & 11.1           & 23.3           & 43.5           & \phantom{0}2.4 & \phantom{0}7.0            & 31.4 & 33.0           & \phantom{0}8.5 \\
\oursbloomsmalllora  & 25.7           & 37.9           & 32.1           & 42.5           & 43.7           & \phantom{0}0.2 & \phantom{0}0.7 & 36.7 & 37.5           & \phantom{0}0.5 \\
\oursbloomsmallfft   & 30.5           & 39.5           & 31.9           & 42.6           & 41.9           & 10.0           & 16.0           & 35.3 & 37.0           & 20.6           \\
\oursvicunasmalllora & 18.6           & \phantom{0}6.4 & \phantom{0}6.0 & 13.5           & 46.6           & \phantom{0}3.5 & \phantom{0}2.1 & 38.3 & 39.7           & 11.6           \\
\oursvicunasmallfft  & 19.4           & 13.4           & 17.1           & 17.4           & 39.6           & \phantom{0}2.9 & \phantom{0}4.4 & 37.8 & 38.1           & \phantom{0}3.6 \\ \bottomrule
\end{tabular}
\caption{Breakdown \dbleu results for the translation tasks from other languages to English.}
\label{tab:dbleu_x_en}
\end{table*}
\begin{table*}[t]
\centering
\small
\begin{tabular}{lcccccccccc}
\toprule
                     & \avgcomet & Ar   & De   & Fr   & It   & Ja   & Ko   & Nl   & Ro   & Zh   \\ \midrule
\multicolumn{11}{l}{\cellcolor{gray!30}\textit{State-of-the-art \sennmt baselines}}                                \\
\nllbsmall           & 72.8      & 76.6 & 63.0 & 79.8 & 72.4 & 74.2 & 76.3 & 68.4 & 67.0 & 77.6 \\
\nllbmedium          & 78.1      & 82.3 & 71.9 & 83.8 & 80.6 & 75.3 & 78.4 & 77.5 & 75.4 & 77.9 \\
\nllbbig             & 78.9      & 82.6 & 76.1 & 83.7 & 80.2 & 74.9 & 80.1 & 78.7 & 76.5 & 77.7 \\
\googletrans         & 81.2      & 81.1 & 82.3 & 84.6 & 84.5 & 75.6 & 76.9 & 84.2 & 83.8 & 78.1 \\ \midrule
\multicolumn{11}{l}{\cellcolor{gray!30}\textit{State-of-the-art LLMs}}                                             \\
\chatgpt             & 85.5      & 85.7 & 86.0 & 87.9 & 88.1 & 81.4 & 82.4 & 87.2 & 87.4 & 83.6 \\
\gptfour             & 86.0      & 86.5 & 86.3 & 88.2 & 88.5 & 81.9 & 83.5 & 87.5 & 87.9 & 84.2 \\ \midrule
\multicolumn{11}{l}{\cellcolor{gray!30}\textit{LLM backbones}}                                                     \\
\model{Llama2}       & 52.2      & 50.3 & 47.1 & 42.6 & 51.6 & 56.1 & 55.7 & 56.5 & 53.0 & 56.9 \\
\model{Bloom}        & 49.4      & 50.6 & 52.8 & 55.5 & 56.8 & 44.5 & 44.3 & 51.2 & 45.2 & 43.3 \\
\model{Vicuna}       & 62.7      & 51.3 & 65.2 & 70.5 & 57.3 & 69.5 & 56.3 & 68.0 & 58.8 & 67.5 \\ \midrule
\multicolumn{11}{l}{\cellcolor{gray!30}\textit{Re-implemented \docnmt baselines}}                                  \\
\docnmtmtfivesmall   & 75.1      & 75.0 & 75.2 & 78.0 & 77.5 & 71.8 & 72.4 & 75.3 & 76.6 & 73.9 \\
\docnmtmtfivebase    & 77.4      & 77.4 & 77.0 & 79.7 & 79.8 & 74.7 & 74.3 & 78.5 & 79.1 & 75.8 \\
\docnmtmtfivelarge   & 78.7      & 79.0 & 78.8 & 80.9 & 80.5 & 75.5 & 75.8 & 80.3 & 80.5 & 76.8 \\
\mrdoctosent         & 79.8      & 80.3 & 79.8 & 82.3 & 81.1 & 76.4 & 76.6 & 81.5 & 81.8 & 78.2 \\
\mrdoctodoc          & ---       & ---  & ---  & ---  & ---  & ---  & ---  & ---  & ---  & ---  \\
\docflat             & 80.3      & 80.2 & 80.0 & 82.5 & 81.7 & 77.4 & 77.6 & 82.2 & 82.3 & 78.4 \\
\iada                & 80.4      & 80.3 & 80.7 & 82.9 & 82.3 & 77.7 & 77.7 & 81.8 & 81.7 & 78.5 \\ \midrule
\multicolumn{11}{l}{\cellcolor{gray!30}\textit{Bilingual English-from/to-Many LLM-based \docnmt models (Ours)}}                         \\
\oursllamasmalllora  & 73.7      & 53.9 & 84.0 & 84.1 & 88.2 & 59.0 & 53.0 & 87.0 & 87.6 & 66.9 \\
\oursllamasmallfft   & 74.0      & 51.6 & 81.9 & 86.5 & 88.3 & 63.6 & 52.9 & 87.0 & 87.3 & 67.1 \\
\oursbloomsmalllora  & 81.4      & 73.3 & 83.6 & 87.0 & 87.1 & 74.0 & 73.8 & 84.8 & 86.0 & 82.6 \\
\oursbloomsmallfft   & 69.9      & 53.7 & 83.2 & 86.9 & 86.8 & 50.5 & 43.3 & 85.1 & 84.4 & 55.5 \\
\oursvicunasmalllora & 71.4      & 54.5 & 82.6 & 87.2 & 64.9 & 57.8 & 53.7 & 87.1 & 87.3 & 67.7 \\
\oursvicunasmallfft  & 77.0      & 77.0 & 81.3 & 87.0 & 88.2 & 65.6 & 55.5 & 84.2 & 87.2 & 67.0 \\ \midrule
\multicolumn{11}{l}{\cellcolor{gray!30}\textit{Multilingual English-from/to-Many LLM-based \docnmt models (Ours)}}              \\
\oursllamasmalllora  & 62.0      & 58.9 & 54.0 & 63.8 & 83.7 & 42.5 & 54.5 & 73.7 & 77.2 & 50.0 \\
\oursllamasmallfft   & 69.6      & 52.9 & 63.2 & 63.2 & 85.1 & 65.3 & 57.0 & 86.3 & 86.3 & 66.8 \\
\oursbloomsmalllora  & 78.7      & 84.0 & 82.5 & 86.5 & 84.6 & 62.4 & 68.4 & 82.3 & 84.2 & 73.8 \\
\oursbloomsmallfft   & 75.8      & 81.0 & 83.3 & 86.2 & 86.9 & 59.1 & 51.7 & 83.9 & 85.7 & 64.4 \\
\oursvicunasmalllora & 60.6      & 52.4 & 58.2 & 57.3 & 78.8 & 43.6 & 47.6 & 83.3 & 83.8 & 40.6 \\
\oursvicunasmallfft  & 65.6      & 50.9 & 47.4 & 53.6 & 87.2 & 61.4 & 56.8 & 83.3 & 86.3 & 63.5 \\ \bottomrule
\end{tabular}
\caption{Breakdown \comet results for the translation tasks from other languages to English.}
\label{tab:comet_x_en}
\end{table*}

\section{Breakdown Results}
\label{sec:breakdown}

We provide detailed breakdowns of the translation tasks from English to other languages, evaluated using \sbleu, \dbleu, and \comet. These are presented in \autoref{tab:sbleu_en_x}, \autoref{tab:dbleu_en_x}, and \autoref{tab:comet_en_x}, respectively. Additionally, we present similar breakdowns for translations from other languages to English, assessed using the same metrics. These results can be found in \autoref{tab:sbleu_x_en}, \autoref{tab:dbleu_x_en}, and \autoref{tab:comet_x_en}.

\begin{table*}[t]
\centering
\small
\begin{tabular}{lcccccccccc}
\toprule
                     & $\mu_{\%}$     & Ar             & De             & Fr             & It             & Ja   & Ko             & Nl             & Ro             & Zh   \\ \midrule
\oursllamasmalllora  & \phantom{0}6.2 & \phantom{0}0.4 & \phantom{0}4.6 & \phantom{0}1.2 & \phantom{0}0.5 & 17.1 & \phantom{0}0.4 & \phantom{0}0.8 & \phantom{0}1.0 & 29.7 \\
\oursllamasmallfft   & \phantom{0}9.4 & \phantom{0}0.3 & \phantom{0}0.9 & \phantom{0}0.3 & \phantom{0}4.4 & 17.7 & \phantom{0}9.4 & 15.4           & \phantom{0}1.2 & 34.7 \\
\oursbloomsmalllora  & 11.2           & \phantom{0}8.4 & \phantom{0}1.0 & 20.8           & \phantom{0}3.9 & 16.4 & \phantom{0}0.0 & \phantom{0}2.8 & \phantom{0}0.9 & 46.9 \\
\oursbloomsmallfft   & 31.8           & 36.6           & 15.8           & \phantom{0}2.7 & 90.1           & 10.7 & \phantom{0}0.1 & 82.0           & \phantom{0}0.2 & 47.7 \\
\oursvicunasmalllora & 10.6           & \phantom{0}0.2 & 15.4           & \phantom{0}0.3 & 13.3           & 15.9 & \phantom{0}0.5 & 20.3           & \phantom{0}0.9 & 28.9 \\
\oursvicunasmallfft  & \phantom{0}8.9 & \phantom{0}0.1 & 14.4           & \phantom{0}0.5 & \phantom{0}0.4 & 27.8 & \phantom{0}0.5 & \phantom{0}4.6 & \phantom{0}0.4 & 31.5 \\ \bottomrule
\end{tabular}
\caption{
Off-target rate (\%) provided by our LLM-based \docnmt models for translation tasks from English to other languages.
$\mu_{\%}$ indicates the average off-target rate.
\textbf{A lower off-target rate indicates better performance}.
}
\label{tab:off_target_en_x}
\end{table*}
\begin{table*}[t]
\centering
\small
\begin{tabular}{lcccccccccc}
\toprule
                     & $\mu_{\%}$     & Ar             & De             & Fr             & It             & Ja           & Ko             & Nl             & Ro             & Zh             \\ \midrule
\oursllamasmalllora  & 29.2           & 87.9           & \phantom{0}2.0 & \phantom{0}4.9 & \phantom{0}1.4 & 25.5         & 44.2           & \phantom{0}1.9 & \phantom{0}1.8 & 93.1           \\
\oursllamasmallfft   & 40.2           & 87.9           & \phantom{0}5.8 & \phantom{0}2.1 & \phantom{0}1.5 & 75.5         & 92.3           & \phantom{0}1.6 & \phantom{0}1.9 & 93.6           \\
\oursbloomsmalllora  & \phantom{0}2.8 & \phantom{0}2.9 & \phantom{0}2.1 & \phantom{0}1.0 & \phantom{0}1.3 & \phantom{0}4.0 & \phantom{0}8.4 & \phantom{0}1.9 & \phantom{0}2.0 & \phantom{0}1.6 \\
\oursbloomsmallfft   & 28.0           & 54.1           & \phantom{0}2.0 & \phantom{0}1.0 & \phantom{0}1.1 & 43.8         & 70.4           & \phantom{0}1.6 & \phantom{0}1.9 & 76.4           \\
\oursvicunasmalllora & 32.3           & 88.2           & \phantom{0}2.6 & \phantom{0}1.2 & 28.0           & 40.4         & 35.7           & \phantom{0}1.9 & \phantom{0}1.9 & 90.5           \\
\oursvicunasmallfft  & 44.7           & 94.1           & \phantom{0}9.0 & \phantom{0}1.3 & \phantom{0}1.3 & 98.3         & 96.6           & \phantom{0}5.3 & \phantom{0}1.9 & 94.6           \\ \bottomrule
\end{tabular}
\caption{
Off-target rate (\%) provided by our LLM-based \docnmt models for translation tasks from other languages to English.
$\mu_{\%}$ indicates the average off-target rate.
\textbf{A lower off-target rate indicates better performance}.
}
\label{tab:off_target_x_en}
\end{table*}
\section{Off-Target Translation}
\label{sec:off_target_app}
We present the complete results on the off-target translation problem in \autoref{tab:off_target_en_x} and \autoref{tab:off_target_x_en}.

\begin{table}[t]
\centering
\small
\setlength{\tabcolsep}{8pt}
\begin{tabular}{lcc}
\toprule
                     & En-De & En-Fr \\ \midrule
\docnmtmtfivelarge   & 39.8  & 26.2  \\ \midrule
\oursllamasmalllora  & \textbf{64.4}  & 29.9  \\
\oursllamasmallfft   & 62.4  & 28.4  \\
\oursbloomsmalllora  & 58.3  & 23.6  \\
\oursbloomsmallfft   & 49.8  & 25.1  \\
\oursvicunasmalllora & 48.9  & \textbf{30.2}  \\
\oursvicunasmallfft  & 43.5  & 27.2  \\ \bottomrule
\end{tabular}
\caption{
Generative accuracy (in \%) on the English-German and English-French contrastive test sets.
Best results are highlighted in \textbf{bold}.
}
\label{tab:gendisco}
\end{table}
\section{Discourse Phenomena}
\label{sec:disco}

\citet{DBLP:journals/corr/abs-2304-12959} propose to evaluate the accuracy on the constrastive test sets in a generative way. Hence, we present the generative accuracy in \autoref{tab:gendisco}.

\begin{figure*}[t]
    \centering
    \begin{subfigure}[t]{0.4\textwidth}
    \centering
    \begin{tikzpicture}[scale=0.6]
        \begin{axis}[
            title={English-Romanian},
            xlabel={Percentage (\%) of training data},
            ylabel={\comet},
            ymin=0, ymax=100,
            symbolic x coords = {
            1, 10, 20, 30, 40, 50, 60, 70, 80, 90, 100
            },
            legend pos=south east,
            legend cell align={left},
            legend style={
                legend columns=2,
                font=\tiny
            },
            width=10cm,
            height=8cm,
            ymajorgrids=true,
            grid style=dashed,
        ]
        
        % l-7b-lora
        \addplot[
            color=blue,
            mark=square,
            ]
            coordinates {
            (1,34.76) 
            (10,79.96) 
            (20,83.87) 
            (30,83.74) 
            (40,84.28)
            (50,84.34)
            (60,84.29)
            (70,84.66)
            (80,83.53)
            (90,84.67)
            (100,83.6)
            };
        \addlegendentry{\oursllamasmalllora}

        % l-7b-fft
        \addplot[
            color=cyan,
            mark=square*,
            ]
            coordinates {
            (1,82.92) 
            (10,80.16) 
            (20,82.37) 
            (30,81.44) 
            (40,82.68)
            (50,83.74)
            (60,81.09)
            (70,83.26)
            (80,80.93)
            (90,84.17)
            (100,82.67)
            };
        \addlegendentry{\oursllamasmallfft}

        % b-7b-lora
        \addplot[
            color=green,
            mark=pentagon,
            ]
            coordinates {
            (1,37.71) 
            (10,63.99) 
            (20,74.05) 
            (30,66.99) 
            (40,69.5)
            (50,65.89)
            (60,74.43)
            (70,62.82)
            (80,68.09)
            (90,69.3)
            (100,61.67)
            };
        \addlegendentry{\oursbloomsmalllora}

        % b-7b-fft
        \addplot[
            color=lime,
            mark=pentagon*,
            ]
            coordinates {
            (1,67.84) 
            (10,63.99) 
            (20,71.05) 
            (30,63.99) 
            (40,60.5)
            (50,67.89)
            (60,70.43)
            (70,72.82)
            (80,65.09)
            (90,68.3)
            (100,71.58)
            };
        \addlegendentry{\oursbloomsmallfft}

        % v-7b-lora
        \addplot[
            color=red,
            mark=diamond,
            ]
            coordinates {
            (1,35.7) 
            (10,80.32) 
            (20,82.5) 
            (30,84.23) 
            (40,83.33)
            (50,81.34)
            (60,83.14)
            (70,83.39)
            (80,82.22)
            (90,84.26)
            (100,85)
            };
        \addlegendentry{\oursvicunasmalllora}

        % v-7b-fft
        \addplot[
            color=magenta,
            mark=diamond*,
            ]
            coordinates {
            (1,75.72) 
            (10,78.32) 
            (20,72.5) 
            (30,74.23) 
            (40,80.33)
            (50,71.34)
            (60,73.14)
            (70,75.39)
            (80,72.22)
            (90,64.26)
            (100,69.64)
            };
        \addlegendentry{\oursvicunasmallfft}

        \end{axis}
        \end{tikzpicture}
        % \caption{English-Romanian}
        \label{fig:scaling_en_ro}
            
    \end{subfigure}
    ~
    \begin{subfigure}[t]{0.4\textwidth}
    \centering
    \begin{tikzpicture}[scale=0.6]
        \begin{axis}[
            title={English-Chinese},
            xlabel={Percentage (\%) of training data},
            ylabel={\comet},
            ymin=0, ymax=100,
            symbolic x coords = {
            1, 10, 20, 30, 40, 50, 60, 70, 80, 90, 100
            },
            legend pos=south east,
            legend cell align={left},
            legend style={
                legend columns=2,
                font=\tiny
            },
            width=10cm,
            height=8cm,
            ymajorgrids=true,
            grid style=dashed,
        ]
        
        % l-7b-lora
        \addplot[
            color=blue,
            mark=square,
            ]
            coordinates {
            (1,34.71) 
            (10,79.55) 
            (20,78.01) 
            (30,78.17) 
            (40,77.93)
            (50,75.17)
            (60,77.07)
            (70,79.62)
            (80,79.04)
            (90,77.53)
            (100,80.64)
            };
        \addlegendentry{\oursllamasmalllora}

        % l-7b-fft
        \addplot[
            color=cyan,
            mark=square*,
            ]
            coordinates { 
            (1,77.06) 
            (10,75.32) 
            (20,77.96) 
            (30,79.79)
            (40,78.06)
            (50,80.45)
            (60,78.48)
            (70,80.62)
            (80,77.33)
            (90,78.27)
            (100,75.48)
            };
        \addlegendentry{\oursllamasmallfft}

        % b-7b-lora
        \addplot[
            color=green,
            mark=pentagon,
            ]
            coordinates {
            (1,50.8) 
            (10,72.71) 
            (20,76.81) 
            (30,69.1) 
            (40,75.11)
            (50,64.11)
            (60,62.23)
            (70,73.36)
            (80,75.91)
            (90,70.32)
            (100,65.51)
            };
        \addlegendentry{\oursbloomsmalllora}

        % b-7b-fft
        \addplot[
            color=lime,
            mark=pentagon*,
            ]
            coordinates {
            (1,66.67) 
            (10,61.11) 
            (20,63.74) 
            (30,74.18) 
            (40,70.74)
            (50,69.66)
            (60,62.91)
            (70,71.82)
            (80,68.33)
            (90,68.65)
            (100,62.23)
            };
        \addlegendentry{\oursbloomsmallfft}

        % v-7b-lora
        \addplot[
            color=red,
            mark=diamond,
            ]
            coordinates {
            (1,38.95) 
            (10,77.6) 
            (20,80.07) 
            (30,81.38) 
            (40,80.67)
            (50,78.92)
            (60,74.36)
            (70,80.91)
            (80,80.95)
            (90,77.12)
            (100,81.31)
            };
        \addlegendentry{\oursvicunasmalllora}

        % v-7b-lora
        \addplot[
            color=magenta,
            mark=diamond*,
            ]
            coordinates {
            (1,79.81) 
            (10,78.9) 
            (20,81.07) 
            (30,80.38) 
            (40,79.67)
            (50,78.12)
            (60,76.36)
            (70,81.91)
            (80,80.05)
            (90,75.12)
            (100,79.44)
            };
        \addlegendentry{\oursvicunasmallfft}

        \end{axis}
        \end{tikzpicture}
        % \caption{English-Chinese}
        \label{fig:scaling_en_zh}
    \end{subfigure}

    \caption{\comet-Percentage (\%) of training data for the translations from English to Romanian, and Chinese.}
    \label{fig:scaling_en_ro_zh}

\end{figure*}
\section{Scaling Law of Parallel Documents from English to Romanian and Chinese}
\label{sec:scaling_en_ro_zh}

In \autoref{sec:analysis}, we find that our LLM-based \docnmt models are highly efficient in terms of the amount of training data. To confirm our findings in \autoref{sec:analysis}, we conduct additional experiments on the translation tasks from English to Romanian and Chinese. As shown in \autoref{fig:scaling_en_ro_zh}, we can confirm the superiority of LLM-based \docnmt models with regard to data efficiency.

\section{Zero-Shot Crosslingual Transfer}
\label{sec:transfer}

\begin{table*}[t]
\centering
\small
\setlength{\tabcolsep}{7pt}
\begin{tabular}{lcccccccccc}
\toprule
                     & $\mu_{\Delta}$   & Ar               & De    & Fr    & It               & Ja               & Ko               & Nl               & Ro              & Zh               \\ \midrule
\oursllamasmalllora  & +29.4            & +36.3            & +38.8 & +37.2 & +32.1            & +15.9            & +17.1            & +21.7            & +35.8           & +29.5            \\
\oursllamasmallfft   & +29.0            & +41.2            & +40.5 & +37.1 & +18.0            & +27.7            & +29.4            & +11.2            & +18.5           & +37.5            \\
\oursbloomsmalllora  & +20.3            & \phantom{0}+7.5  & +40.7 & +20.7 & +21.9            & +17.5            & +15.9            & +23.7            & +25.3           & \phantom{0}+9.8  \\
\oursbloomsmallfft   & +27.3            & +14.8            & +37.8 & +28.9 & +43.3            & +13.1            & +15.3            & +38.5            & +34.7           & +19.5            \\
\oursvicunasmalllora & \phantom{0.}-8.9 & \phantom{.}-12.6 & +22.1 & +18.9 & \phantom{.}-28.6 & \phantom{.}-27.8 & \phantom{.}-18.7 & \phantom{.}-11.8 & +12.1           & \phantom{.}-34.1 \\
\oursvicunasmallfft  & \phantom{0.}-1.4 & \phantom{0}+7.3  & +25.2 & +17.7 & \phantom{.}-14.6 & \phantom{.}-24.7 & \phantom{0.}-5.3 & \phantom{.}-21.8 & \phantom{0}+7.6 & \phantom{0.}-3.5 \\ \bottomrule
\end{tabular}
\caption{
    The difference ($\Delta$) in \comet scores on the test sets from English to other languages between our English-German LLM-based \docnmt models and their backbones. $\mu_{\Delta}$ indicates the average difference across all the languages in this table.
}
\label{tab:zero_shot}
\end{table*}

We also explore the transferability of translation capabilities acquired from one language pair to others. We assess our English-German LLM-based \docnmt models on English-to-other-language test sets, comparing their \comet scores to their base models in \autoref{tab:zero_shot}. Our results indicate that models with fine-tuned instructions (\model{Llama2-7B} and \model{Bloom-7B}) consistently exhibit positive transfer effects across all language pairs, while those with instruction-tuned models (\model{Vicuna-7B}) benefits only few languages. 

% We present the complete results on zero-shot crosslingual transfer in \autoref{tab:zero_shot}.

\end{document}